\PassOptionsToPackage{normalem}{ulem}
\documentclass[]{arxiv_template}
\usepackage{enumitem}
\usepackage[toc,page,header]{appendix}
\usepackage{float}

\usepackage{xcolor}
\definecolor{BoxBorder}{HTML}{C5D4E3} % 边框浅蓝灰色
\definecolor{BoxBadge}{HTML}{648CB4}  % 标题栏的背景蓝色

\newtcolorbox{appendixbox}[1]{
    enhanced,
    breakable,
    title=#1,              % 传入的标题
    colback=white,         % 框内背景色设为纯白
    colframe=BoxBorder,    % 边框颜色
    boxrule=1.5pt,         % 边框粗细
    arc=4pt,               % 边框圆角半径
    drop fuzzy shadow=gray!40, % 添加右下角柔和阴影
    attach boxed title to top left={xshift=12pt, yshift=-\tcboxedtitleheight/2},
    boxed title style={
        colback=BoxBadge,  % 标题栏背景色
        colframe=BoxBadge, % 标题栏边框色与背景一致
        arc=3pt,           % 标题栏圆角
        boxrule=0pt,
        top=3pt, bottom=3pt, left=8pt, right=8pt % 标题栏内边距
    },
    coltitle=white,                 % 标题字体颜色为白色
    fonttitle=\sffamily\bfseries,   % 标题字体加粗
    left=15pt, right=15pt, top=15pt, bottom=15pt % 框内文本的外边距
}

\usepackage{listings}
\definecolor{JsonString}{HTML}{2B5D9A} % 字符串的颜色（深蓝）
\definecolor{JsonBackground}{HTML}{F8F9FA} % 代码块背景色（浅灰）

\lstdefinelanguage{json}{
    basicstyle=\small\ttfamily, % 字体大小及等宽字体
    showstringspaces=false,     % 不显示字符串中的空格标记
    breaklines=true,            % 允许自动换行
    backgroundcolor=\color{JsonBackground}, % 背景色
    frame=single,               % 加上细线边框
    rulecolor=\color{black!20}, % 边框颜色
    stringstyle=\color{JsonString}, % 字符串颜色
    morestring=[b]",            % 识别双引号里的内容为字符串
}

\usepackage{textcomp}
\definecolor{orange-web}{RGB}{255, 165, 0}
\definecolor{sagegreen}{RGB}{120, 150, 120}
\usepackage{makecell}
\usepackage{graphicx}
\usepackage{algorithm}
\usepackage{tabularx}
\usepackage{multirow}
\usepackage{pifont}
\usepackage{amsmath,amssymb,amsthm}
\usepackage{etoolbox}
\usepackage{lipsum}
\usepackage{tablefootnote}
\usepackage{threeparttable}
\usepackage{wrapfig}
\usepackage{colortbl}
\usepackage{xspace}
\usepackage{longtable}
\usepackage{listings}
\usepackage{fontawesome5}
\definecolor{boxback}{gray}{0.95}
\definecolor{promptback}{gray}{0.98}
\definecolor{outputback}{RGB}{230, 245, 255}
\definecolor{titleblue}{RGB}{0, 110, 180}

\usepackage{algorithm}
\usepackage{algpseudocode}
\usepackage{amsmath}

\usepackage{CJKutf8}

\newcommand{\hi}[1]{\vspace{.25em} \noindent {\bf #1} }
\newcommand{\llm}{\textsc{LLM}\xspace}
\newcommand{\llms}{\textsc{LLMs}\xspace}
\newcommand{\bfit}[1]{\textbf{\textit{#1}}}
\newcommand{\oursys}{\textbf{\textsf{Workspace-Bench}}\xspace}
\newcommand{\colearn}{{Workspace Learning}\xspace}

\newcommand{\zw}[1]{\textcolor{purple}{#1}}

\newcommand{\zxh}[1]{\textcolor{red}{#1}}
\newcommand{\tzr}[1]{\textcolor{orange}{#1}}
\newcommand{\addcontent}[1]{\textcolor{blue}{[XH: #1]}}

\title{\oursys1.0: Benchmarking AI Agents on Workspace Tasks\\ with Large-Scale File Dependencies}

\makeatletter
\gdef\authorlist{%
  \begin{center}
    \sffamily\normalsize
    Zirui Tang$^{1,2*}$,
    Xuanhe Zhou$^{1*\dagger}$,
    Yumou Liu$^{1}$,
    Linchun Li$^{1}$,
    Yukai Wu$^{1}$,
    Weizheng Wang$^{1}$,
    Hongzhang Huang$^{1}$,
    
    Wei Zhou$^{1}$,
    Jun Zhou$^{1}$,
    Jiachen Song$^{1}$,
    Shaoli Yu$^{1}$,
    Jinqi Wang$^{1}$,
    Zihang Zhou$^{1}$,
    Hongyi Zhou$^{1}$,
    
    Yuting Lv$^{2}$,
    Jinyang Li$^{5}$,
    Jiashuo Liu$^{5}$,
    Ruoyu Chen$^{2}$,
    Chunwei Liu$^{3}$,
    GuoLiang Li$^{4}$,
    Jihua Kang$^{2\dagger}$,
    Fan Wu$^{1}$
  \end{center}
}
\makeatother

\makeatletter
\gdef\affiliationlist{%
  \begin{center}
    \affiliationfont\sffamily
    $^{1}$Shanghai Jiao Tong University
    $^{2}$ByteDance
    $^{3}$MIT
    $^{4}$Tsinghua University
    $^{5}$Independent Researher
  \end{center}
}
\makeatother

\abstract{
Workspace learning requires AI agents to identify, reason over, exploit, and update explicit and implicit dependencies among heterogeneous files in a worker's workspace, enabling them to complete both routine and advanced tasks effectively. Despite its importance, existing relevant benchmarks largely evaluate agents on pre-specified or synthesized files with limited real-world dependencies, leaving workspace-level evaluation underexplored. To this end, we introduce \oursys, a benchmark for evaluating AI agents on \textbf{Workspace Learning} involving Large-Scale File Dependencies. We construct realistic workspaces with 5 worker profiles, 74 file types, 20,476 files (up to 20GB) and curate 388 tasks, each with its own file dependency graph, evaluated across 7,399 total rubrics that require cross-file retrieval, contextual reasoning, and adaptive decision-making. We further provide \oursys-Lite, a 100-task subset that preserves the benchmark distribution while reducing evaluation costs by about 70\%.
We evaluate 4 popular agent harnesses and 7 foundation models. Experimental results show that current agents remain far from reliable workspace learning, where the best reaches only $\sim$60\%, substantially below the human+tool result of 80.7\%, and the average performance across agents is only 43.3\%.
}

\titleformat*{\paragraph}{\bfseries}
\checkdata[Project Page]{\url{https://workspace-bench.github.io/}}
\checkdata[HuggingFace]{\url{https://huggingface.co/datasets/Workspace-Bench}}
\checkdata[GitHub Repository]{\url{https://github.com/OpenDataBox/Workspace-Bench}}

\begin{document}
\begin{CJK*}{UTF8}{gbsn}

\maketitle

\tcbset{reset}

{
\renewcommand{\thefootnote}{\fnsymbol{footnote}}
  \footnotetext[1]{Equal Contribution} 
  \footnotetext[2]{Corresponding author: Xuanhe Zhou, Jihua Kang}% 
}

\begin{figure}[h]
    \vspace{-1em}
    \centering
    \includegraphics[width=0.95\textwidth]{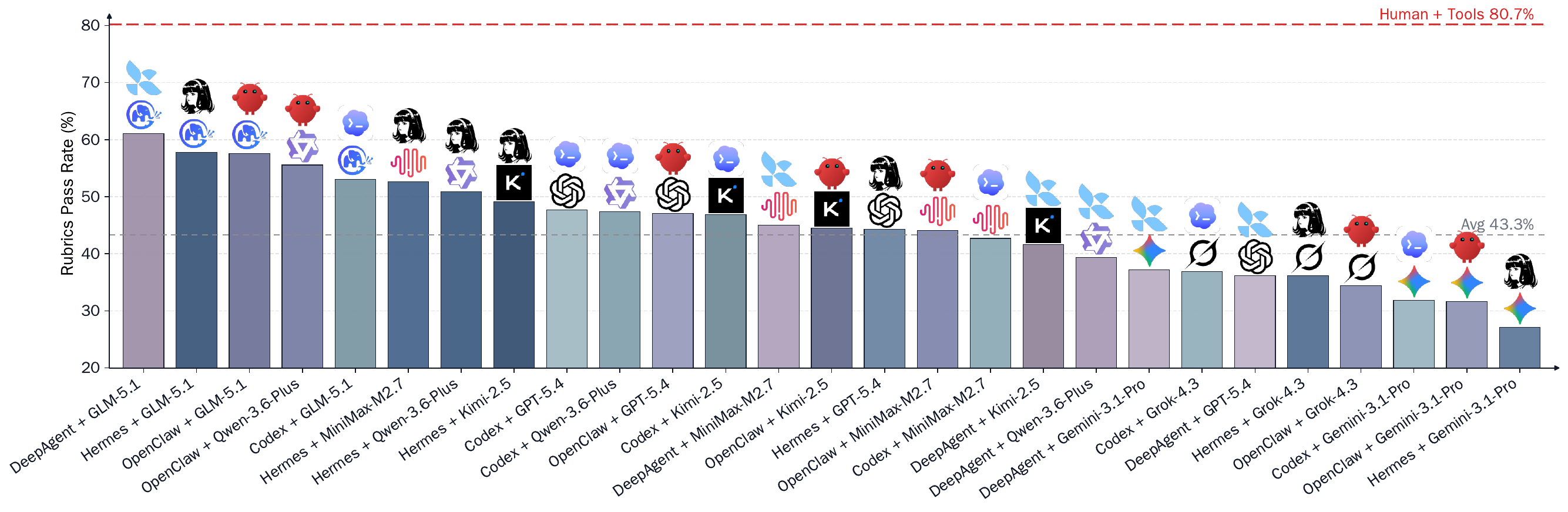}
    \vspace{-0.4cm}
    \caption{Performance on the lite version of \oursys, covering combinations of agent harnesses (DeepAgent, Hermes, and OpenClaw) with different backbone LLMs. See up-to-date comparisons results on our project homepage.} %The top three configurations all pair GLM-5.1 with different harnesses, while lower-ranked combinations exhibit substantial variation across both harnesses and LLMs.
    
    %     \caption{Performance on \oursys-Lite. We evaluate four harnesses (ClaudeCode, DeepAgent, Hermes, OpenClaw) and seven backbone LLMs. Configurations pairing Opus-4.7 with OpenClaw, ClaudeCode, and Hermes dominate the top three, which suggests that the high success rates are largely attributable to the robust capabilities of Opus-4.7, highlighting a lack of specialized optimization for \colearn within current agent harnesses.}
    \label{fig:rubrics_result}
\end{figure}

% 图表说明：上图展示了从依赖参数化记忆的 Pre-trained Model，到利用情景化记忆的 Context Learning，再到我们提出的、依赖关系化记忆的 \colearn 的范式演进。这一演进过程反映了模型为应对日益增长的任务复杂度和真实世界对齐度，其能力和自主性的必然发展方向。

\begin{table*}[b]
\centering
\caption{Comparison of Agent Benchmarks Across Six Workspace Character Dimensions.}
\renewcommand{\arraystretch}{1.2}
% 定义绿色勾选指令
\newcommand{\cmark}{\textcolor{green!70!blue}{\ding{51}}}
% 定义红色叉号指令
\newcommand{\xmark}{\textcolor{red}{\ding{56}}}
\label{tab:intro-bench-compare}
\resizebox{\textwidth}{!}{
\begin{tabular}{l|c|cccccc|c}
\hline
\textbf{Benchmark} & \textbf{\begin{tabular}[c]{@{}c@{}}Input\\ Modality\end{tabular}} & \textbf{\begin{tabular}[c]{@{}c@{}}Workspace\\ Structures\end{tabular}} & \textbf{\begin{tabular}[c]{@{}c@{}}File\\ Modalities\end{tabular}} & \textbf{\begin{tabular}[c]{@{}c@{}}Task-Supporting\\ Files\end{tabular}} & \textbf{\begin{tabular}[c]{@{}c@{}}Result-Providing\\ Files\end{tabular}} & \textbf{\begin{tabular}[c]{@{}c@{}}Semantic Content\\ Relations\end{tabular}} & \textbf{\begin{tabular}[c]{@{}c@{}}File Lineage\\ Relations\end{tabular}} & \textbf{\begin{tabular}[c]{@{}c@{}}Labeling Time\end{tabular}} \\ 
\hline
OneMillion-Bench~\cite{OneMillionBench} & P & \xmark & \xmark & \xmark & \xmark & \xmark & \xmark & Over 2000H   \\
CL-Bench~\cite{dou2026clbench} & P & \xmark & \xmark & \xmark & \xmark & \xmark & \xmark &    \\
\hline
Odysseys Bench~\cite{odysseys} & T & \xmark & \xmark & \xmark & \xmark & \xmark & \xmark & --   \\
OSWorld~\cite{osworld} & T & \xmark & \xmark & \xmark & \xmark & \xmark & \xmark & Approximately 1800H   \\
MMTB~\cite{MMTB} & T & \xmark & \xmark & \xmark & \xmark & \xmark & \xmark & --   \\
MultiAgentBench~\cite{zhu2025multiagentbench} & T & \xmark & \xmark & \xmark & \xmark & \xmark & \xmark &  --   \\
CRMArena-Pro~\cite{crmarena} & T & \xmark & \xmark & \xmark & \xmark & \xmark & \xmark & --   \\
\hline
OfficeQA-Pro~\cite{OfficeQA} & F\&T & \xmark & \xmark & \cmark & \cmark & \cmark & \xmark & -- \\ 
GDPVal~\cite{gdpval} & F\&T & \xmark & <10 & \cmark & \cmark & \cmark & \xmark & --   \\
\hline
SWE-Bench~\cite{jimenez2023swebench} & F\&T & Single & <5 & \cmark & \cmark & \xmark & \xmark & --  \\
WorkBench~\cite{workbench} & F\&T & Single & <5 & \cmark & \xmark & \cmark & \xmark & --  \\
OfficeBench~\cite{wang2024officebench} & F\&T & Single & <10 & \cmark & \cmark & \cmark & \xmark & --   \\
TheAgentCompany~\cite{xu2024theagentcompany} & F\&T & Single & <20 & \cmark & \cmark & \cmark & \xmark & Approximately 3000H   \\
\hline
\textbf{Ours (\oursys)} & F\&T & Multi & >70 & \cmark & \cmark & \cmark & \cmark & Over 2500H\\ \hline
\end{tabular}
}
\caption*{Input Modality:  P (Prompt-only), T (Tool-based), F (File-based).}
\end{table*}

\section{Introduction}
\label{sec:introduction}

% The Rise of AI Agents. Acknowledge the rapid development of fully automated agents (e.g., OpenHands, Anthropic Computer Use) and their potential to revolutionize office work.

Recent advances in foundation models and agent harnesses have substantially expanded the operational scope of AI agents. Beyond model inference, these agents provide system-level capabilities for connecting to external tools through MCP and skills, maintaining task state and long-term memory, orchestrating multi-step execution, enforcing guardrails, and supporting systematic evaluation mechanism~\cite{microsoft2026copilot,hermes,openclaw2026}. 
These capabilities make AI agents increasingly useful for reducing human effort in many daily and even advanced tasks~\cite{xi2023}. However, completing real-world workplace tasks (e.g., sales plan updates, email-based interview scheduling) requires these agents to move beyond isolated capabilities and engage in \textit{\textbf{\colearn}}: the ability to connect tasks with relevant data and understand lineage and logical relationships across numerous workspace files (see Section~\ref{sec:problem}). This requires agents to identify, reason over, exploit, and update explicit and implicit dependencies across heterogeneous files, enabling them to navigate complex digital environments effectively.

% Developing practical AI agents (assistants) that can handle real-world workplace tasks over numerous heterogeneous and multimodal files remains challenging.
% Recent advances in foundation models and agent harnesses have substantially expanded the operational scope of AI agents.  
% Beyond model inference, these agents provide system-level capabilities for connecting to external tools through MCP and skills, maintaining task state and long-term memory, orchestrating multi-step execution, enforcing guardrails, and supporting systematic evaluation~\cite{microsoft2026copilot,anthropic2026cowork,hermes,openclaw2026}. 
% These capabilities make AI agents increasingly useful for reducing human effort in daily and advanced workplace tasks, such as cross-file information consolidation, context-critical spreadsheet construction, and routine business workflow execution.
% \zxh{-- Sync 1. harness techniques 2. workplace tasks with fig2}

% including closed-source systems such as Copilot~\cite{microsoft2026copilot} and Cowork~\cite{anthropic2026cowork}, as well as open-source frameworks such as OpenClaw~\cite{openclaw2026} and Hermes~\cite{hermes}

%However, \textit{their ability to solve real-world office problems remains highly questioned.}

However, a persistent gap remains between the apparent capabilities of current AI agents and their actual performance on real-world workplace tasks~\cite{kwa2025metr,gartner2025agentic}. 
On one hand, many specialized professional workflows (e.g., cross-departmental financial reconciliation, compliance-sensitive report generation) are difficult and costly to delegate directly to AI agents. For instance, 49\% of enterprises identify inference cost as the top blocker for scaling AI agents, with nearly half spending 76--100\% of their AI budget on inference alone~\cite{digitalocean2026currents}. 
%When an agent fails to understand the underlying structure of a task, it often resorts to brute-force exploration through iterative reasoning loops, consuming up to 1{,}000$\times$ more tokens than non-agentic interactions~\cite{bai2026tokenspend}, with per-session costs reaching \$6 for a single coding task~\cite{vantage2026agenticcost}. 
On the other hand, even on simplified analogues of such workplace tasks in existing benchmarks, the most advanced agents still perform poorly. For instance, the best-performing AI agent achieves only 24--30\% task completion in TheAgentCompany~\cite{xu2024theagentcompany}; and 47\% on multi-application office workflows in OfficeBench~\cite{wang2024officebench}. 

We conducted an in-depth analysis of 154 authentic task scenarios sourced from the Lark platform in ByteDance. The investigation reveals that, while AI agents excel at overcoming surface-level tasks, such as navigating complex Graphical User Interfaces (GUIs) and executing multi-turn tool invocations, they still struggle severely when interacting with massive, fragmented document workspaces. 
For instance, in commercial settings, drafting a highly tailored proposal requires multi-file coordination across unstructured client profiles, historical communication records, and structured internal industry knowledge bases. Completing such tasks often requires navigating dozens of content-related files, where existing agents frequently struggle, leading to critical information omissions, logical inconsistencies, and factual inaccuracies. 

Thus, there is an urgent need for a benchmark that can thoroughly test the above capabilities on real-world workplace tasks. However, as shown in Table~\ref{tab:intro-bench-compare}, existing benchmarks fail to effectively simulate authentic office workflows and complex inter-file relationships. 
%Specifically, many benchmarks decouple task execution from processing and reasoning over an actual digital workspace with numerous files, or merely provide task-specific, pre-packaged files to the agent. As a result, they lack a holistic directory structure where agents must independently search and filter information. Although some benchmarks, representing the closest attempts to simulating complete file systems that require dynamic tool invocation and reasoning, do construct a workspace structure, they still exhibit critical bottlenecks in reflecting the full complexity of real-world scenarios.
Specifically, \textbf{Prompt-Driven} benchmarks (e.g., OneMillion-Bench~\cite{OneMillionBench}, CL-Bench~\cite{dou2026clbench}), which embed all requisite information entirely within natural language instructions, and \textbf{Open-Source-Driven} benchmarks (e.g., Odysseys Bench~\cite{odysseys}, CRMArena-Pro~\cite{crmarena}), which require agents to depend on tool usage to query web or API environments without upfront data, both fundamentally bypass the core medium of daily office workflows: processing and reasoning over actual digital workspace with numerous files. \textbf{Task-File-Driven} benchmarks (e.g., OfficeQA-Pro~\cite{OfficeQA}, GDPVal~\cite{gdpval}) introduce file handling by providing task-specific, pre-packaged files to the agent. However, they resemble QA over independent files, and lack a holistic directory structure where agents must independently search and filter information. \textbf{Workspace-Relevant} benchmarks (e.g., OfficeBench~\cite{wang2024officebench}, TheAgentCompany~\cite{xu2024theagentcompany}, ClawsBench~\cite{clawsbench}, ClawMark~\cite{clawmark}) represent the closest attempts to simulating complete file systems that require dynamic tool invocation and reasoning. Nevertheless, they still exhibit critical bottlenecks in reflecting the full complexity of real-world scenarios. 
First, they rely on monolithic, single-style file system structures, lacking persona-dependent diversity. Second, they predominantly cover fewer than 10 basic file modalities (e.g., \textit{xlsx, docx, pdf}), missing more than 50 diverse formats typically encountered in real office scenarios. More importantly, while existing tasks may inherently involve multiple files, they generally treat inter-file synergies as implicit byproducts rather than explicitly evaluating task-to-data dependency identification, failing to consider aspects like (1) aggregating result-providing files, (2) reasoning over semantic content relations, and (3) comprehending contextual task-supporting files. Crucially, they entirely omit file lineage relations, which are vital to reflect the agents' ability to trace version histories and derivations. % , which is a hallmark of dynamic, authentic office environments

\begin{figure}[!t]
    \centering
    \includegraphics[width=\textwidth]{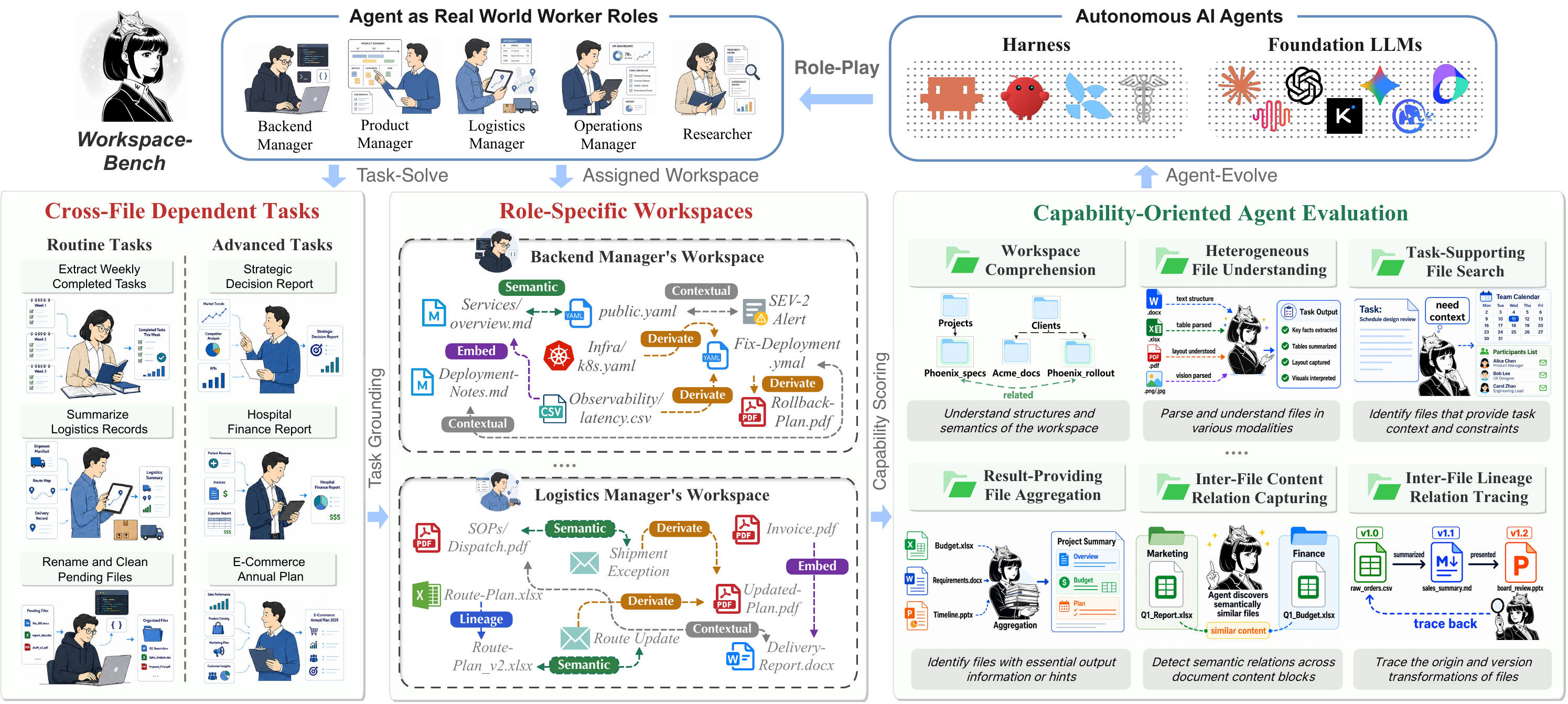}
    %\caption{Panorama of Office Tasks That Necessitate \colearn.}
    \caption{Overview of \oursys.}
    \label{fig:intro-examples}
    % : Integrating multi-source enterprise data into actionable deliverables
\end{figure}

To address this critical evaluation gap, we introduce \oursys, a benchmark designed to systematically measure an agent's \colearn capabilities. \oursys is built around three core principles. (1) \oursys provides a realistic environment composed of five distinct user profiles, including an operations manager, a logistics manager, a product manager, a backend developer, and a researcher, each with a file ecosystem of total 20,476 interconnected files, chats, and artifacts (up to 20GB) that mirror the complex digital workspace of a real knowledge worker. (2) \oursys includes over 388 file-dependency-driven tasks with 7399 rubrics designed to probe the six evaluation dimensions across multiple difficulty levels, ranging from basic file organization to cross-functional report generation. (3) \oursys offers a fine-grained evaluation testbed in which each task is paired with a set of rubrics (19.1 in average) that assess not only the correctness of the final output but also critical intermediate decisions.

% To address this critical evaluation gap, we introduce \oursys, a benchmark designed to systematically measure an agent's \colearn capabilities. \oursys is built around three core principles.  (1) \oursys provides a \textbf{realistic environment} composed of five distinct user profiles-, including an operations manager, a logistics manager, an AI product manager, an researcher, and a backend developer, where each with a file ecosystem of thousands of interconnected files, chats, and artifacts that mirror the complex and often disordered digital workspace of a real knowledge worker. (2) \oursys includes over \tzr{XXX} \textbf{dependency-driven tasks} with \tzr{XXX} rubrics designed to probe the \tzr{six dimensions and ten types} of \colearn (Table~\ref{tab:intro-bench-compare}) across multiple difficulty levels, ranging from basic file organization to cross-functional report generation, and requiring multi-step reasoning such as locating relevant files, extracting key information, and synthesizing results into shared records.  (3) \oursys offers a \textbf{fine-grained evaluation} testbed in which each task is paired with a set of rubrics that assesses not only the correctness of the final output (e.g., whether a summary contains the required project risks), but also critical intermediate decisions (e.g., whether the agent selected the correct file version or identified the appropriate data source). Furthermore, rubrics difficulty and necessity are also explicitly labeled as core dimensions of the evaluation.

\textbf{Benchmark Impact.} Through \oursys, we aim to shift the evaluation of AI assistants and fully-automated AI agents from isolated skills toward workspace-aware reasoning. Our empirical results show that, despite impressive progress in foundation models, state-of-the-art agents still struggle significantly when faced with tasks that require genuine \colearn. For instance, across 27 combinations of 4 agent harnesses and 7 backbone LLMs, the average Rubrics Pass Rate is merely 43.3\%. The best-performing combination (DeepAgent + GLM-5.1) only achieves nearly 60\% accuracy. Furthermore, we observe massive performance gaps between different agent harnesses, with open-source solutions like DeepAgent + MiniMax-M2.7 struggling with severe ``cost explosions'', which consume up to 58.1 interaction turns and 0.61 million tokens per task while still failing to achieve competitive success rates (averaging only 45\% pass rate). This highlights a fundamental and underexplored bottleneck on the path from capable language models to truly reliable productivity agents.

Our contributions are summarized as follows:

\noindent$\bullet$ We propose \oursys, a benchmark for evaluating workspace tasks involving large-scale file dependencies. It contains five realistic user workspaces, heterogeneous files, and 388 file-dependency-driven tasks, enabling systematic evaluation of agents' \colearn capability. %reasoning over complicated workspace structures.

\noindent$\bullet$ We develop a \colearn evaluation framework, combining sandboxed parallel execution, automatic workspace recovery, and dependency-graph-based assessment. %to measure both task success and workspace dependency recognition.
It enables fine-grained assessment over 7,000+ rubrics, covering correctness, intermediate reasoning, operational efficiency.

\noindent$\bullet$ We evaluate 15 configurations utilizing state-of-the-art 
foundation models and agent harnesses on \oursys. The results reveal a clear 
performance deficit on workspace tasks: 
(1) agents suffer consistent degradation from Easy (57.6\%) to Hard (40.5\%) 
workspace tasks; (2) heterogeneous file understanding and lineage tracing are the primary capability bottlenecks across all agent configurations; 
(3) current harnesses show limited impact on powerful models but serve as effective performance boosters for weaker foundation models in dependency-aware task solving; and (4) human-agent collaboration (80.7\%) still significantly outperforms fully autonomous execution.

\section{Related Work}
\label{sec:related}

\subsection{Automated Agent Techniques}

\hi{GUI and Desktop Agents.}
Recent advancements in multimodal Large Language Models (LLMs) have spurred the development of agents capable of directly interacting with Graphical User Interfaces (GUIs). Early works like SeeClick~\cite{seeclick} and CogAgent~\cite{cogagent} focused on improving GUI grounding---the ability to map natural language instructions to specific pixel coordinates or UI elements on a screen. More recently, systems such as UFO~\cite{ufo} and ShowUI~\cite{showui} have demonstrated the ability to execute multi-step operations within Windows or mobile OS environments. Foundation models specifically trained for GUI tasks, such as UI-TARS~\cite{uitars}, have further pushed the boundaries of what agents can achieve without relying on underlying DOM trees or accessibility APIs. Commercial products,such as Microsoft Copilot Cowork~\cite{microsoft2026copilot} and Perplexity Computer~\cite{perplexity2026computer}, now deploy these techniques to function as general-purpose desktop assistants. However, while these agents excel at localized, single-application operations, they often struggle when tasks require understanding the implicit relationships between scattered data sources across a complex file system.

\hi{Memory and RAG for Agents.}
To handle long-horizon tasks and extensive context, modern agents heavily rely on Retrieval-Augmented Generation (RAG)~\cite{lewis2020rag} and persistent memory architectures. Systems like MemGPT~\cite{packer2024memgpt} manage memory hierarchically, allowing agents to retain user preferences and past interactions across sessions. While these techniques expand the volume of accessible information, they typically treat retrieved context as a flat collection of text chunks. They lack the native ability to model the structural and temporal dependencies between these chunks (such as version lineage or role constraints) which is the core focus of \colearn in \oursys.

\subsection{Agent Benchmarks}
To systematically evaluate the capabilities of LLM-based agents, numerous benchmarks have emerged. Based on their information dependency and environment interaction, existing efforts can be broadly categorized into four paradigms.

\hi{Prompt-Driven Benchmarks.}
These benchmarks embed all requisite task information entirely within natural language instructions, focusing on an agent's reasoning and comprehension capabilities under information-complete conditions. For instance, CL-Bench~\cite{dou2026clbench} evaluates Context Learning by requiring agents to learn new rules from provided text. Similarly, OneMillion-Bench~\cite{OneMillionBench} offers a massive scale of instruction-following tasks across economically consequential scenarios. While critical for evaluating pure reasoning, these benchmarks require zero interaction with external environments or actual digital files, fundamentally bypassing the operational core of office workflows.

\hi{Open-Source/Environment-Driven Benchmarks.}
To evaluate proactive information gathering and execution, this paradigm requires agents to heavily depend on tool usage to interact with dynamic environments (e.g., APIs, the Web, or operating systems). Because no upfront data is provided, agents must autonomously invoke tools to acquire the necessary task information. OSWorld~\cite{osworld} and GAIA~\cite{gaia} construct comprehensive, multi-application operating system environments to design open-ended tasks. With a stronger emphasis on visual interfaces, ScreenSpot-Pro~\cite{screenspotpro} and WindowsAgentArena~\cite{windowsagentarena} specifically evaluate an agent's GUI interaction and visual grounding capabilities. Shifting from desktop to browser-based execution, WebArena~\cite{webarena} and Odysseys Bench~\cite{odysseys} focus on complex web navigation and cross-website task completion. Meanwhile, from a data-centric perspective, benchmarks like CRMArena-Pro~\cite{crmarena} and MultiAgentBench~\cite{zhu2025multiagentbench} are built upon data sources, requiring agents to iteratively invoke relevant tools to explore, query, and retrieve information.
Although these benchmarks successfully incorporate multi-step execution, they predominantly focus on action grounding or API orchestration. Consequently, they largely ignore the fundamental medium of daily knowledge work: the navigation, reasoning, and management within complex, relational local file ecosystems.

\hi{Task-File-Driven Benchmarks.}
Moving closer to real-world data processing, benchmarks in this category introduce actual file handling to evaluate document comprehension and analysis. For example, OfficeQA-Pro~\cite{OfficeQA} grounds its evaluation in enterprise document workflows by providing necessary source text files and reference documents alongside the tasks. Similarly, GDPVal~\cite{gdpval} requires agents to complete specific tasks and generate outputs based on supplied reference files. Expanding beyond pure text, DataCross~\cite{datacross2026} proposes a benchmark for unified, insight-driven analysis across heterogeneous modalities. However, despite incorporating real digital files, these benchmarks treat tasks in isolation by directly feeding task-specific, pre-packaged files to the agent. This approach resembles isolated Document QA rather than authentic office work. Consequently, agents are entirely spared from the realistic challenge of independently searching, filtering, and discovering essential information from a complex file ecosystem.

\hi{Workspace-Relevant Benchmarks.}
Representing the closest approximations to reality, these benchmarks simulate a complete work structure requiring dynamic tool invocation. WorkBench~\cite{workbench} provides tasks based on 5 databases, yet represents them solely as \textit{.xlsx} files, effectively bypassing the complexities of both database systems and hierarchical file navigation. OfficeBench~\cite{wang2024officebench} constructs a file system based on common office file formats, while SWE-bench~\cite{jimenez2023swebench} anchor their evaluations within real-world code repositories. TheAgentCompany~\cite{xu2024theagentcompany} further simulates a corporate cloud environment on OneDrive to test multi-application workflows. Nevertheless, despite their advances, they collectively fall short of replicating the complexity of authentic scenarios. Structurally, they are limited to a single style of file system (e.g., generic office folders or pure codebases) and lack the diversity of personas and organizational contexts. In terms of content coverage, they typically support a few basic file formats, missing the rich tapestry encountered in real knowledge work. More critically, from a task design perspective, many challenges can be resolved by focusing on a single file, thereby failing to compel the agent to reason across the deep, relational dependencies that characterize real office work. Consequently, they lack systematic evaluation for essential inter-file synergies.

In contrast, \oursys is explicitly designed to target the core gap: the relational structure of a single agent's knowledge workspace. It moves beyond static file provision to systematically evaluate the comprehensive dimensions of workspace reasoning. This is achieved by incorporating diverse user personas, supporting over 70 file modalities, and, most importantly, by constructing tasks that necessitate understanding and navigating the intricate web of semantic, aggregative, and lineage-based relations among files.

\section{Collection and Curation of \oursys} 
\label{sec:framework}

% •	目标：详细描述SYNERGON数据集的构建流程，确保其科学性和可复现性。
    % #-Environments, #-tasks, #-files-per-env, #-dependency-levels, #-relation-types, #-dependencies-per-task

% 1.1	Step 1: User Persona Design：定义4种用户画像（产品经理、研究员、工程师、市场专员），并说明其典型的文件操作行为模式。

% 1.2	Step 2: Environment Construction：
    % •	文件系统结构：描述workspace/下的目录结构（projects, personal, shared, archive等）。
    % •	文件来源与类型：说明文件来源于真实开源数据集（如Enron邮件集）和真实工作文档，并给出文件类型分布（引用《知识管理AgentBenchmark.pdf》中的表格）。
    % •	注入冗余与歧义：说明如何通过添加相似文件、过时版本、命名歧义来增加依赖解析的难度。

% 1.3	Step 3: Task Generation and Annotation：
    % •	任务来源：说明任务基于真实办公场景、其他Benchmark（Part2数据）和标注者经验生成。
    % •	标注流程：描述标注者如何根据《协同Benchmark任务标注.pdf》的规范，为每个任务编写自然语言指令、定义评估Rubrics，并最关键地，标注出完成该任务所需的完整依赖图谱 G。

% 1.4	Step 4: Quality Validation：通过交叉验证和试评测来确保任务描述的清晰性、依赖图谱的完备性和评估准则的客观性。

% 1.5	Step 5: Temporal Evolution Simulation：描述如何模拟用户随时间推移的操作（创建、修改、删除、移动文件），并生成每日的系统状态快照和事件日志。

%\addcontent{仔细补充详细步骤和细节;考虑补充一个收集的数据源、血缘版本关系等的table}

To evaluate \colearn beyond static and isolated task settings, we develop \oursys, a benchmark built around realistic digital workspaces and context-grounded office tasks. \oursys is designed to assess whether an agent can operate over heterogeneous files, {divers workspace structures}, and implicit organizational context, different from many other benchmarks adopting a clean collection of independent files. To ensure both realism and reproducibility, we construct \oursys through a controlled pipeline that combines persona-driven workspace simulation, hybrid file collection and generation, task curation, dependency annotation, and expert validation. % Instead of releasing a clean collection of independent files, \oursys simulates workspaces where the information is fragmented, redundant, temporally evolving, and linked through explicit or implicit dependencies. 

% Instead of releasing a fixed snapshot of files, we simulate the lifecycle of knowledge work and assess agents on reasoning over heterogeneous data sources, temporal changes, and organizational context. 

% (1) persona-driven workspace scaffolding, (2) workspace population, (3) temporal evolution simulation, (4) task curation, and (5) quality assurance.

\subsection{Design Principles}

We design \oursys according to four principles that distinguish it from existing agent and document benchmarks.

\hi{High-Fidelity Relational Workspaces.}
Existing benchmarks often place data in clean and independent files, whereas real workplace tasks require agents to navigate messy digital workspaces. Information is typically distributed across folders, modalities, versions, and organizational roles. Therefore, \textit{\oursys aims to construct realistic workspaces with thousands of interconnected artifacts, where agents must account for implicit conventions, role-specific file organization, and noisy workspace structures.}

% \oursys simulates this complexity by constructing a dense environment featuring five distinct professional personas (e.g., PMs, Developers, Ops Managers). %\paragraph{$\star$ Advantage of \oursys:} Unlike static benchmarks, \oursys populates these workspaces with thousands of \bfit{interconnected artifacts}, including chat threads, email attachments, and multi-versioned documents. By mirroring the disordered reality of a professional workspace, it forces agents to handle implicit knowledge (e.g., naming conventions) and role constraints, testing their ability to function not just as calculators, but as integrated \bfit{team members} who understand the ``who, what, and where'' of an organization.

\hi{Dependency-Driven Reasoning.}
Many cross-file benchmarks primarily test surface-level aggregation. In practice, workspace tasks often require retrieving contextually related files from different locations and reasoning over their dependencies (e.g., explicit references, semantic relations, modality transformations, version lineage). Thus, \textit{\oursys aims to explicitly annotate and evaluate dependency-driven interactions among files, rather than treating each file as an isolated evidence source.}

% \oursys introduces over 240 tasks specifically designed to probe the six dimensions of Associative Data Learning. These tasks are not solvable through simple keyword search; they require tracing the lineage of a decision across a project's lifecycle.

%\paragraph{$\star$ Advantage of \oursys:} Our tasks are built on deep dependencies. For example, an agent might need to reconcile a specific code comment with a decision made in a meeting note, while filtering out obsolete information from ``report\_v1'' in favor of ``report\_final\_edited''. This requires temporal awareness and cross-modal synthesis, the ability to link arguments across heterogeneous sources (CSV, Doc, Chat) to form a coherent understanding of the current project state.

\hi{Authentic Task Annotation.}
LLM-generated tasks can scale rapidly, but they often miss the structural complexity and implicit constraints of real professional workflows, especially when the tasks require navigation over multimodal and interdependent workspaces. \textit{\oursys therefore aims to curate tasks from real office scenarios and annotate them manually with domain experts. LLMs are used only as auxiliary tools for verification and rubric optimization, while task logic, dependency specification, and reference outputs remain human-curated.}

%\paragraph{$\star$ Advantage of \oursys:} Rather than synthesizing artificial workflows, we curate task exemplars directly from real-world professional environments. These tasks undergo comprehensive manual annotation by human experts to guarantee practical relevance. While agents are employed in an auxiliary capacity for verifying task solvability and edge cases, the core logic and ground truths are entirely driven by human intelligence. Finally, stringent manual refinement guarantees that every task within \oursys authentically reflects the rigorous demands of human knowledge work.

\hi{Process-Aware Fine-Grained Evaluation.}
A single success-rate score is insufficient for diagnosing agent behavior in workspace tasks. For example, an agent may produce a plausible final summary while relying on an obsolete file version or ignoring a required supporting document. \textit{\oursys therefore aims to evaluate not only final outputs, but also intermediate decisions, including whether the agent identifies the correct files, respects dependency constraints, and uses the appropriate file versions.}

%\paragraph{$\star$ Advantage of \oursys:} \oursys provides a structured evaluation testbed with fine-grained rubrics. We do not just score the final output, but also assess critical intermediate decisions, such as whether the agent identified the correct data source among noisy distractors or respected version lineage. This multi-level scoring ensures that a high mark on \oursys represents true reasoning and reliability, rather than a lucky ``hallucination'' that happens to match the ground truth.

\begin{figure*}[!t]
    \centering
    \includegraphics[width=.95\textwidth]{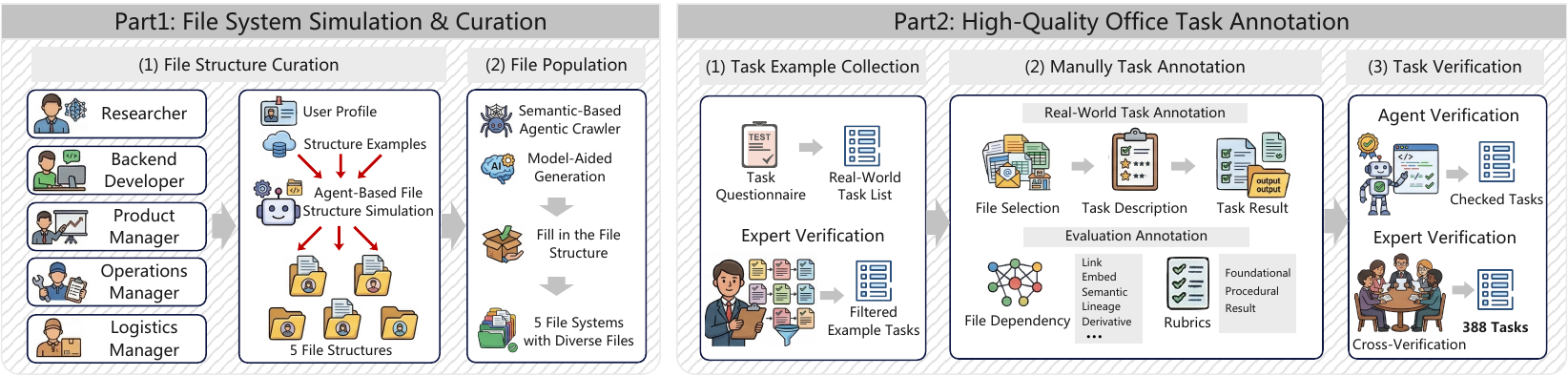}
    \caption{The data collection and curation pipeline of \oursys.}
    \label{fig:bench_prepare}
\end{figure*}

\subsection{Workspace Construction}

We construct workspaces for five representative professional roles in an {internet company}: \textit{Operations Manager}, \textit{Logistics Manager}, \textit{Product Manager}, \textit{Backend Developer}, and \textit{Researcher}~\cite{filesys-roles}. These roles cover diverse workspace structures and corresponding tasks.

As existing agent evaluations are often conducted in cleaned sandbox environments, which differ substantially from real digital workspaces. 
In practice, a workspace usually evolves in a top-down manner: users first establish a workflow-aligned directory hierarchy and then populate it with downloaded resources, authored documents, intermediate drafts, and derived artifacts. This process naturally produces three properties: (1) deeply nested directory structures, (2) semantically noisy files such as obsolete drafts and historical revisions, and (3) implicit cross-file dependencies. To simulate these properties, we design a top-down, two-stage hybrid construction pipeline (see Figure~\ref{fig:bench_prepare}).

\hi{Structure Generation.}
Different professional roles organize files according to different workflows. For instance, developers commonly maintain directories such as \texttt{src/}, \texttt{tests/}, and \texttt{docs/}, while researchers may organize files around \texttt{literature/}, \texttt{experiments/}, and \texttt{results/}. To capture this diversity, we first define detailed persona profiles for the five roles, including their responsibilities, typical workflows, file usage patterns, and domain-specific terminology. Conditioned on these profiles, we prompt agents to generate tree-structured directory hierarchies that reflect role-specific workspace organization. We also introduce controlled structural noise (e.g., redundant folders, ambiguous names, archive directories) to better match real-world file systems.

\hi{Content Population.}
After constructing the directory, we populate each workspace using a hybrid strategy that combines real-world data retrieval and grounded generation. We first deploy a semantic-driven agentic crawler that traverses the generated directory tree and retrieves public resources relevant to the semantic workspace directory, such as arXiv papers, GitHub repositories, technical documents, reports, spreadsheets, and presentation materials. We then use LLMs to synthesize related artifacts grounded in the collected files, such as emails discussing a paper, meeting notes referring to a design document, or reports derived from spreadsheets. %This process enriches the workspace with realistic file relationships while avoiding the homogeneity of purely synthetic corpora.

%\hi{Content Population.} Following the establishment of the directory framework, typical user operations entail two synergistic activities, including retrieving authentic external data or subsequently authoring related documents (e.g., analysis reports or discussion emails). To emulate this process, we employ a hybrid strategy combining real-world data retrieval with LLM generation. Specifically, we first deploy a semantics-driven Agentic Crawler that traverses the directory tree, autonomously scraping highly relevant, authentic public resources (e.g., intricately formatted arXiv papers, real-world GitHub repositories) based on the semantic context of the file paths. Subsequently, LLMs are leveraged to synthesize related content grounded in the collected data (e.g., generating an email discussing a specific paper or a memo analyzing a codebase), which injects deep file relationships throughout the workspace.

% This strategy effectively mitigates the homogenization inherent in purely generated text, successfully injecting Deep File Relationships throughout the extensive file system.

Finally, domain experts review the simulated file systems to verify their plausibility and structural consistency, focusing on whether the directory hierarchy matches the intended persona, whether file contents are coherent with their locations, and whether injected file relationships can support meaningful workspace tasks.

The resulting workspaces embed several major challenges for agent evaluation. First, \emph{task-related file retrieval} requires agents to navigate nested directory structures and identify relevant files from noisy candidates. Second, \emph{lineage understanding} requires agents to distinguish file relations like understanding multiple revisions of the same artifact, such as \texttt{report\_v1}, \texttt{report\_reviewed}, and \texttt{report\_final}. Third, \emph{heterogeneous-source reasoning} requires agents to connect information across modalities, such as linking a slide chart to its source spreadsheet or connecting a discussion email to the corresponding design document.

\subsection{Workspace Task Curation}

Based on the constructed workspaces, we curate \textbf{388} tasks for \oursys. Each task is written as a natural language request and is intentionally under-specified: an agent must inspect the workspace structure and recover the relevant file dependencies to complete the task. The tasks cover both routine operations, such as form filling and file organization, and complex multi-step requests, such as preparing weekly reports by reconciling prior documents, recent code changes, and project status updates. 
% These tasks are designed to evaluate capabilities including \emph{workspace exploration}, \emph{contextual retrieval}, \emph{cross-directory discovery}, \emph{dependency tracing}, and \emph{heterogeneous file understanding}.
To ensure realism and evaluability, we avoid LLM-based fully automated task generation and instead adopt a problem-driven human curation pipeline (see Figure~\ref{fig:bench_prepare}).

\hi{Sourcing Authentic Workflows.} We first collect real workplace workflows through an internal questionnaire, where participants provide task descriptions, expected inputs, and desired outputs. Domain experts then filter the collected workflows to remove trivial or underspecified cases and retain tasks that require nontrivial workspace exploration and cross-file reasoning. The selected workflows are standardized into 154 representative task scenarios, such as synthesizing client profiles, purchase histories, and interaction logs to derive customer scores and personalized recommendations.

%. Domain experts then filtered these submissions to isolate high-value, complex cases, which are standardized into foundational exemplars. Ultimately, this process yielded 154 representative task scenarios (e.g., synthesizing client profiles, purchase histories, and interaction logs to formulate customer scores and tailored product recommendations).

\hi{Multi-dimensional Task Annotation.} Starting from these representative scenarios, 25 human annotators aligned with the five workspace roles create concrete tasks within the simulated workspaces. For each task, annotators write the natural language instruction, identify the required inputs, produce a reference output, and design evaluation rubrics. Since many tasks have open-ended outputs, we use rubric-based evaluation instead of relying on a single exact-match answer. Each rubric consists of fine-grained binary propositions that assess output quality, procedural correctness, and task completion. These propositions are grouped into foundational, procedural, and result-oriented criteria.
%Starting from these representative scenarios, 25 human annotators aligned with the five workspace roles create concrete tasks within the simulated environments. As the highly open-ended nature of these outputs precludes precise evaluation via a single ground-truth answer, we introduce rubric-based assessment. These rubrics comprise multiple fine-grained binary propositions (e.g., ``Does the estimated time account for 5 minutes?'') designed to verify output quality, and are categorized into foundational, procedural, and result-oriented types. Furthermore, to precisely quantify the agent's \colearn capabilities, annotators constructed a file dependency graph for each task, which is defined as ``the minimal set of essential file paths'' an agent must access to successfully resolve the request. Finally, the annotators explicitly tagged the required capability dimensions and calibrated task difficulty (Simple, Medium, Hard), which is rigorously anchored to the \colearn capability hierarchy, where tasks requiring Level 1-2 capabilities are classified as Simple, those strictly dependent on Level 2 are designated as Medium, and those necessitating advanced Level 3 capabilities are categorized as Hard (See Figure~\ref{fig:paradigm}).

Annotators further construct a file dependency graph for each task. The graph specifies the minimal set of essential file paths that an agent must access or use to solve the task correctly. This annotation enables process-aware evaluation of whether an agent has discovered the right evidence, used the correct file versions, and followed the required dependency structure. Annotators also tag each task with the required capability dimensions and assign a difficulty level. 
Following the six core challenges mentioned in Figure~\ref{fig:intro-examples}, tasks only requiring workspace exploration and result-providing files utilization are labeled as \textit{Easy}, tasks primarily requires semantic content relations understanding and task-supporting files utilization are labeled as \textit{Medium}, and tasks with heterogeneous file understanding and lineage understanding are labeled as \textit{Hard}.

% Following the \colearn capability hierarchy, tasks requiring Level 1--2 capabilities are labeled as \textit{Simple}, tasks primarily dependent on Level 2 capabilities are labeled as \textit{Medium}, and tasks requiring advanced Level 3 capabilities are labeled as \textit{Hard} (see Figure~\ref{fig:paradigm}).

\hi{Ensuring Objective Evaluability.} Open-ended tasks often introduce ambiguity in human-written rubrics. To improve objectivity, we use an auxiliary agent pipeline to convert vague criteria into data-grounded assertions. For example, a criterion such as ``Is the calculation correct?'' is converted into a verifiable assertion such as ``Does the final value equal [specific value]?''. Human experts then cross-validate all tasks, dependency graphs, reference outputs, and rubrics to ensure annotation consistency and evaluation reliability. This process yields a curated benchmark of 388 tasks with explicit workspace dependencies and fine-grained evaluation criteria.

\section{Benchmark Analysis}
\label{sec:benchmark}

% 我们现在有如下数据维度：
% 1. 每个任务的协同类别（可能有多个类别）
% 2. 每个任务的难度 （简单，中等，困难）
% 3. 每个任务需要的能力维度（6个维度，可能有多个维度）
% 4. 每个文件系统所具有的任务数量
% 5. 每个任务具有的Rubrics数量的个数
% 6. 每一条Rubrics的难度（简单、中等、困难）
% 7. 每一条Rubrics的类型（基础、过程、结果）
% 8. 每一条Rubrics的必要性（必要的，不必要的，可选的）

\begin{figure*}[!t]
    \centering
    \includegraphics[width=.99\textwidth]{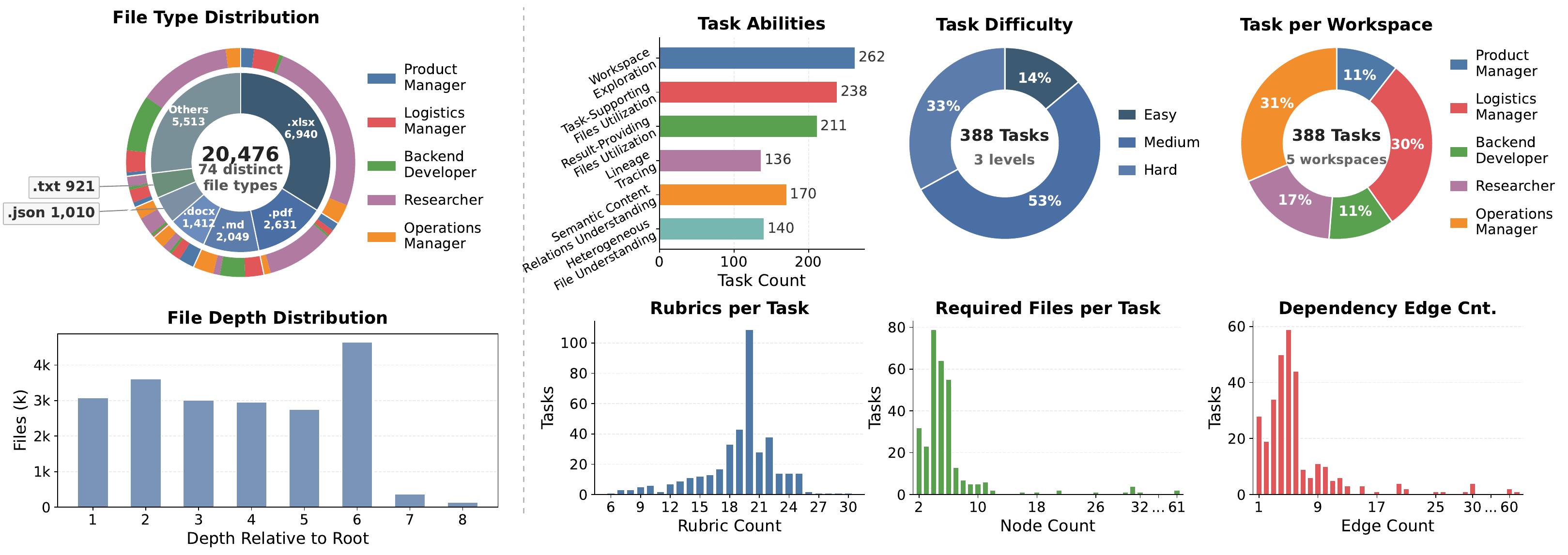}
    \caption{Workspace (left) and task (right) distribution of the \oursys.}
    \label{fig:bench_statistics}
\end{figure*}

In this section, we present a comprehensive statistical overview of the \oursys benchmark. To systematically evaluate \colearn, we constructed five large-scale and realistic digital workspaces. We analyze the benchmark from multiple dimensions, including the overall scale, workspace composition, task distribution, and dependency complexity, to demonstrate its quality, diversity, and alignment with real-world challenges.

\subsection{Overall Statistics}

% (clbench) \#-context, tasks, rubrics | tasks per context, rubrics per task, input length (token) | examples

% (futureX) domain coverage, event types, dynamic (Volatility), different tiers (difficulty)

\oursys consists of 388 tasks distributed across \textbf{five} distinct professional workspaces, each embodying a realistic job persona: \textit{Operations Manager} (122 tasks), \textit{Logistics Manager} (115 tasks), \textit{Researcher} (67 tasks), \textit{Backend Developer} (43 tasks), and \textit{Product Manager} (41 tasks). Unlike traditional benchmarks that provide a handful of isolated files per task, each workspace in \oursys is a large-scale, self-contained digital environment containing up to 11,020 files, with an average of over 4,000 files per workspace. Across all five workspaces, the benchmark encompasses 74 heterogeneous file types,
% spanning office documents (\texttt{.docx}, \texttt{.pdf}), spreadsheets (\texttt{.xlsx}, \texttt{.csv}), presentations (\texttt{.pptx}), source code (\texttt{.java}, \texttt{.py}, \texttt{.js}), configuration files (\texttt{.yaml}, \texttt{.json}), statistical data (\texttt{.dat}, \texttt{.sav}), emails (\texttt{.eml}), and multimedia (\texttt{.png}, \texttt{.wav}), 
reflecting the full heterogeneity of real-world digital workspaces (see top-left of Figure~\ref{fig:bench_statistics}). Furthermore, task outputs span more than 16 distinct file formats, including analytical reports, spreadsheets, presentations, and scripts, mirroring the diverse deliverables expected in professional settings.

To ensure rigorous and multi-dimensional evaluation, each task is annotated with a comprehensive dependency graph and evaluated using fine-grained rubrics. On average, a task requires resolving 5.1 dependency edges across 4.7 different files, and its execution is assessed against 19.1 rubric items (ranging from 6 to 30 per task). Table~\ref{tab:overall_stats} summarizes the key statistics of \oursys.

\begin{table}[h]
\centering
\caption{Overall Statistics of \oursys. The benchmark features large-scale realistic workspaces, complex dependency graphs, and rigorous fine-grained evaluation rubrics.}
\label{tab:overall_stats}
\resizebox{0.9\textwidth}{!}{%
\begin{tabular}{@{}lc|lc@{}}
\toprule
\textbf{Metric} & \textbf{Value} & \textbf{Metric} & \textbf{Value} \\ \midrule
Total Workspaces (Personas) & 5 & Avg. Files per Workspace & 4,095 \\
Total Tasks & 388 & Max Files in a Workspace & 11,020 \\
Total Files Across Workspaces & 20,476 & Total Directories & 3，299 \\
% Total Annotated Dependency Edges & 1,486 & Avg. Dependency Edges per Task & 3.8 \\
Total Evaluation Rubrics & 7,399 & Avg. Rubrics per Task & 19.1 \\
File Types Covered & 74 & Max. Directory Depth & 8 \\
Output File Formats & 16 & Avg. Expert Annotation Hours / Task & >3h \\ \bottomrule
\end{tabular}%
}
\end{table}

\subsection{Workspace Composition and Heterogeneity}

A core feature of \oursys is its realistic workspace population, which mirrors the messy and heterogeneous nature of real digital environments. As illustrated in Figure~\ref{fig:bench_statistics}, the files within each workspace span 74 modalities and formats, including documents (e.g., \texttt{.docx}, \texttt{.pdf}, \texttt{.md}), spreadsheets (\texttt{.xlsx}, \texttt{.csv}), presentations (\texttt{.pptx}), code repositories (\texttt{.java}, \texttt{.py}, \texttt{.js}, \texttt{.ts}), configuration files (\texttt{.yaml}, \texttt{.json}), emails (\texttt{.eml}), statistical datasets (\texttt{.dat}, \texttt{.sav}, \texttt{.xpt}), and images (\texttt{.png}, \texttt{.jpg}).

Specifically, spreadsheets and documents constitute the two dominant categories, accounting for 37.5\% and 35.3\% of the total files, respectively, reflecting their prevalence in professional office scenarios. Code and configuration files contribute a further 12.7\%, driven primarily by the \textit{Backend Developer} workspace, which alone contains 43 distinct file extensions. The five workspaces exhibit markedly different compositions, where the \textit{Researcher} workspace is the largest with 11,020 files spread across 2,059 directories, while the \textit{AI Product Manager} workspace is the most compact with 1,379 files organized in 309 directories.
To simulate the temporal evolution of real workspaces, a subset of the files have multiple historical versions (e.g., \texttt{v1}, \texttt{v2}, \texttt{final}), requiring agents to reason over \textit{Lineage Tracing}. Furthermore, the files are deeply nested within hierarchical directory structures, with an average file-parent depth of 3.7 and a maximum depth of 8, forcing agents to actively navigate and discover information rather than relying on flat retrieval.

\subsection{Task Distribution and Dependency Complexity}

\begin{figure*}[!t]
    \centering
    \includegraphics[width=.99\textwidth]{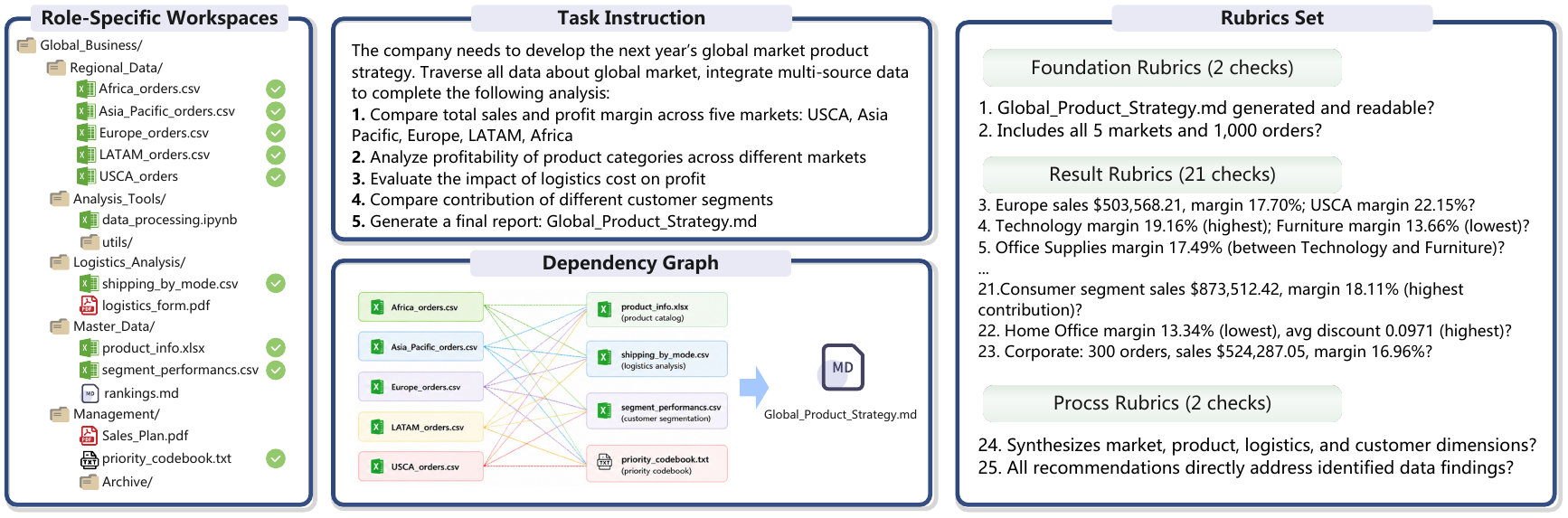}
    \caption{An illustrative task example from \oursys.}
    \label{fig:case_example}
\end{figure*}

The tasks in \oursys are carefully curated to cover the six dimensions of \colearn, as summarized in the \textit{Task Abilities} panel of Figure~\ref{fig:bench_statistics}. Since one task can require multiple abilities, the counts are multi-label task assignments rather than a partition of the 388 tasks. \textit{Workspace Exploration} is the most frequent ability, appearing in 262 tasks (67.5\%), indicating that agents often need to inspect directory structures before locating relevant evidence. \textit{Task-Supporting Files Utilization} appears in 238 tasks (61.3\%), requiring agents to infer and use files that provide necessary task context rather than relying only on explicit paths. \textit{Result-Providing Files Utilization} appears in 211 tasks (54.4\%), reflecting the need to consolidate evidence across files that directly support the final deliverable. The remaining dimensions further stress realistic workspace reasoning: \textit{Content Relations Understanding} appears in 170 tasks (43.8\%), \textit{Semantic Heterogeneous File Understanding} in 140 tasks (36.1\%), and \textit{Lineage Tracing} in 136 tasks (35.1\%).

Figure~\ref{fig:bench_statistics} also shows that the benchmark spans all five workspaces and three difficulty levels. The task-per-workspace panel shows broad coverage across \textit{Operations Manager} (122 tasks, 31\%), \textit{Logistics Manager} (115 tasks, 30\%), \textit{Researcher} (67 tasks, 17\%), \textit{Backend Developer} (43 tasks, 11\%), and \textit{Product Manager} (41 tasks, 11\%). The task-difficulty panel further shows that most tasks are \textit{Medium} difficulty (53\%), followed by \textit{Hard} (33\%) and \textit{Easy} (14\%), where difficulty is determined by the number of required execution steps and the complexity of the underlying collaboration types.

Moreover, the complexity of \oursys is reflected in its dependency graphs. We categorize tasks into three difficulty levels based on the number of annotated dependency edges in \texttt{file\_deps} (each directed link between files counts as one edge). The first two levels span equal-width integer ranges of edge counts, which yields a balanced partition of tasks:

\begin{itemize}
    \item \textbf{Low Edge Density (0--2 edges):} Accounting for 33.8\% of tasks, these instances involve only a handful of explicit file-to-file dependencies, typically corresponding to light cross-file aggregation or a small set of direct references.
    \item \textbf{Moderate Edge Density (3--5 edges):} Accounting for 36.9\% of tasks, these instances connect several artifacts at once, requiring the agent to maintain consistency across multiple sources (e.g., aligning figures across a report, a spreadsheet, and supplementary notes).
    \item \textbf{High Edge Density ($\ge$6 edges):} Accounting for 29.4\% of tasks, these instances exhibit a dense dependency web among many files, demanding broader coordination and more careful propagation of information before producing the final deliverable.
\end{itemize}

To enable fine-grained and multi-faceted evaluation, each task is annotated with an average of 19.1 rubrics. The 7,399 rubrics are categorized into three types: \textit{Result-oriented} rubrics (54.8\%) verify the correctness and completeness of the final output; \textit{Foundation} rubrics (25.0\%) check basic task compliance such as file naming, format, and storage location; and \textit{Process-oriented} rubrics (20.2\%) assess whether the agent follows a sound reasoning and execution process. 

Figure~\ref{fig:case_example} illustrates a representative Operations Manager task from \oursys, requiring generating a global market product strategy. This complex workflow requires the agent to explore the workspace, identify 9 core files, and synthesize multi-dimensional data across markets, products, and logistics. The final output is evaluated against a strict set of 25 rubrics, which are divided into 2 foundation, 21 result, and 2 process checks. This setup effectively tests the agent's capacity for dependency-aware reasoning and cross-document aggregation in a realistic office environment.

% In terms of difficulty, rubrics span three levels---\textit{Easy} (28.2\%), \textit{Moderate} (49.5\%), and \textit{Hard} (20.5\%)---with a small fraction of expert-level rubrics (1.7\%) that test deep domain knowledge and nuanced multi-step verification.

\subsection{\oursys-Lite}

In addition to the full benchmark, we introduce \textbf{\oursys-Lite}, a curated subset of \textbf{100 tasks} specifically designed for lightweight and rapid evaluation. By selecting across all five workspaces, three difficulty levels, and six \colearn dimensions, this subset strictly preserves the distributional fidelity of the original dataset. Consequently, \textbf{\oursys-Lite} delivers a robust and comprehensive evaluation while reducing evaluation costs by approximately 70\%.

\begin{figure*}[!t]
    \centering
    \includegraphics[width=.99\textwidth]{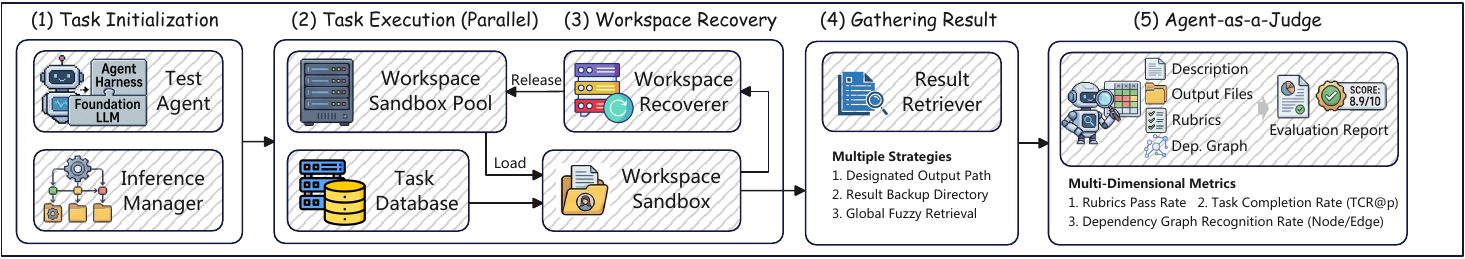}
    \caption{Evaluation Framework of \oursys.}
    \label{fig:evaluation_framework}
\end{figure*}

\subsection{Evaluation Framework}

% 简单介绍我们的评估框架motivation，以及大致模块与流程。 (1) 介绍评估系统的组成部分 (2) 介绍使用的指标 (3) 介绍评估流程：一个任务会经历什么具体流程最终得到评估结果。

To evaluate agent performance on realistic workspace-grounded tasks, we designed a comprehensive evaluation framework (see Figure~\ref{fig:evaluation_framework}).

\hi{Agent Initialization and Task Execution.} Initially, the execution configurations for each agent are statically declared via a unified YAML file. Upon parsing these configurations, the Inference Manager schedules the corresponding test tasks from the task suite and provisions an isolated workspace sandbox from the Sandbox Pool, specifically tailored to the target user profile. Subsequently, the agent is deployed into this isolated environment and driven by prompts to execute the specific task.

\hi{Parallel Evaluation.} To overcome the efficiency bottlenecks associated with large-scale tasks, we introduce a dual parallel acceleration mechanism, operating at both the workspace and task levels. First, the workspace-level parallelism leverages five independent user profile workspace, which schedule and execute different tasks concurrently within their corresponding isolated workspaces, yielding a 5x evaluation throughput. Second, concerning task-level parallelism, we pre-clones multiple image replicas of identical workspace and integrates them into a Sandbox Pool for dynamic management. Upon the dispatch of a new evaluation task, the scheduling engine automatically allocates an idle sandbox environment for execution, which achieves a more pronounced acceleration in the overall evaluation process.

\hi{Task Result Collection.} Following task completion, agents frequently store their output files in unpredictable, non-fixed paths within the workspace. To accurately capture these target artifacts from a massive volume of files, we employ a Multi-strategy File Extraction Technique, which incorporates three parallel retrieval mechanisms.
\textbf{(1) Instruction-constrained path extraction.} During the task assignment phase, the system's prompt compels the agent to explicitly state the exact path of the result file in its final response, enabling direct file retrieval.
\textbf{{(2) Unified replica-based centralized retrieval.}} The system enforces the agent to save an additional copy of the result in a globally designated directory, allowing the evaluation framework to directly scan and extract the output.
\textbf{(3) Metadata-based global fuzzy matching.} Utilizing target file characteristics (e.g., expected filenames) defined in the task metadata, the system executes a comprehensive traversal and fuzzy search across the entire sandbox file system.
Finally, the system aggregates deduplicated file lists acquired through these three concurrent strategies for subsequent evaluation.

\hi{Workspace Recovery.} Once task result extraction is complete, the modified sandbox workspace must be reset to its initial state. The system independently maintains a standard baseline workspace snapshot for each user profile. Upon task completion, a parallel recursive algorithm is employed to compute the directory tree discrepancies between the current workspace and the baseline. As shown in Alg.~\ref{alg:bfs_rollback}, it compares node states layer by layer starting from the root, which replicates missing nodes from the baseline, forcibly deletes extraneous nodes, and overwrites modified nodes that exhibit mismatched binary hashes with their baseline counterparts. This mechanism ensures the rapid roll back of the manipulated environment, after which the sandbox is released back into the pool for subsequent task reuse.

\begin{algorithm}[t]
\caption{Parallel BFS Workspace Rollback}
\label{alg:bfs_rollback}
\begin{algorithmic}[1]
\Require $MRoot$: root path of manipulated sandbox, $SRoot$: root path of standard workspace, $MaxP$: maximum concurrency limit
\Ensure Restored $MRoot$ matching $SRoot$

\State $Layer \gets \{(MRoot, SRoot)\}$ \Comment{Initialize BFS with root nodes}
\While{$Layer \neq \emptyset$}
    \State $NextLayer \gets \emptyset$
    
    \Statex \Comment{Process directories concurrently, bounded by $MaxP$}
    \ForAll{$(MDir, SDir) \in Layer$ \textbf{in parallel}}
        \ForAll{$node \in MDir \cup SDir$}
            \If{$node \in MDir \setminus SDir$}
                \State \Call{Delete}{$MDir.node$} \Comment{Delete extraneous nodes}
            \ElsIf{$node \in SDir \setminus MDir$}
                \State \Call{Copy}{$SDir.node \to MDir$} \Comment{Add missing nodes}
            \ElsIf{\Call{IsModified}{$MDir.node, SDir.node$}}
                \State \Call{Replace}{$MDir.node$ \textbf{with} $SDir.node$} \Comment{Replace files}
            \ElsIf{\Call{IsDirectory}{$node$}}
                \State $NextLayer \gets NextLayer \cup \{(MDir.node, SDir.node)\}$
            \EndIf
        \EndFor
    \EndFor
    
    \State $Layer \gets NextLayer$ \Comment{Proceed to the next BFS depth}
\EndWhile
\end{algorithmic}
\end{algorithm}

\hi{Agent-as-a-Judge.} For the final assessment, we employ an Agent-as-a-Judge paradigm. To evaluate rubric satisfaction, the judge agent is provided with the deduplicated output files, original inputs fils, task rubrics, and the evaluated agent's execution trajectory. Based on these inputs, it generates binary correctness scores, confidence metrics, detailed rationales, and categorized error types. Furthermore, to compute the dependency recognition rates, the judge agent first dynamically extracts a predicted dependency graph from the execution trajectory, and subsequently compares it against the predefined ground-truth graph.

\hi{Evaluation Metrics.} We employ a combination of established and novel metrics to comprehensively assess agent performance in terms of execution correctness, reasoning capabilities, and operational efficiency.

\textbf{(1) Rubric Pass Rate (\%).} This metric calculates the overall proportion of correctly satisfied evaluation rubrics across all tasks. Formally, it is defined as $R_{acc} = \frac{N_{passed}}{N_{total}}$, where $N_{passed}$ is the number of successfully met rubrics and $N_{total}$ is the total number of evaluated rubrics.

\textbf{(2) Task Completion Rate (\%).} This denotes the percentage of tasks in which the agent successfully satisfies a threshold proportion of the task-specific rubrics. For instance, TCR$@80$ represents the fraction of tasks where the agent meets at least 80\% of the required rubrics. It is mathematically formulated as $\text{TCR}@80 = \frac{1}{|T|} \sum_{i=1}^{|T|} \mathbb{I}(s_i \geq p)$, where $T$ represents the total set of tasks, $s_i$ is the rubric completion ratio for the $i$-th task, and $\mathbb{I}(\cdot)$ is the indicator function.

\textbf{(3) Dependency Graph Recognition Rate (\%).} To comprehensively evaluate the structural accuracy of the file dependency graph dynamically extracted from the agent's execution trajectory, we compare the predicted graph against a predefined ground-truth graph at both the node and edge levels using F1-Score. Specifically, for the node set, we define node precision as $NP = \frac{|V_{pred} \cap V_{gt}|}{|V_{pred}|}$ and node recall as $NR = \frac{|V_{pred} \cap V_{gt}|}{|V_{gt}|}$, where $V_{pred}$ and $V_{gt}$ denote the vertex sets of the predicted and ground-truth graphs, respectively. The corresponding \textbf{Node F1 Score} is computed as $NF1 = \frac{2 \cdot NP \cdot NR}{NP + NR}$. Similarly, for the edge set, we define edge precision as $EP = \frac{|E_{pred} \cap E_{gt}|}{|E_{pred}|}$ and Edge Recall as $ER = \frac{|E_{pred} \cap E_{gt}|}{|E_{gt}|}$, where $E_{pred}$ and $E_{gt}$ represent the predicted and ground-truth edge sets. The corresponding \textbf{Edge F1 Score} is given by $EF1 = \frac{2 \cdot EP \cdot ER}{EP + ER}$. 

% \textbf{(3) Dependency Graph Recognition Rates (\%).} To comprehensively evaluate the structural accuracy of the file dependency graph dynamically extracted from the agent's execution trajectory, we measure its overlap with a predefined ground-truth graph at both the node and edge levels. Specifically, the \textbf{Node Recognition Rate (NRR)} assesses task-required file identification accuracy, defined as $NRR = \frac{|V_{pred} \cap V_{gt}|}{|V_{gt}|}$, where $V_{pred}$ and $V_{gt}$ denote the vertex sets of the predicted and ground-truth graphs, respectively. Building upon this, the \textbf{Edge Recognition Rate (ERR)} quantifies the accuracy of the identified relationships, formulated as $ERR = \frac{|E_{pred} \cap E_{gt}|}{|E_{gt}|}$, where $E_{pred}$ and $E_{gt}$ represent the corresponding edge sets.

\textbf{(4) Average Token Consumption (tokens).} This measures the mean total number of tokens, combining both input prompts and output completions, consumed to complete a single task. It is expressed as $\bar{C}_{token} = \frac{1}{|T|} \sum_{i=1}^{|T|} c_i$, where $c_i$ is the total token count for the $i$-th task.

\textbf{(5) Average Task Turns.} This metric reflects the average number of interaction steps, tool invocations, or reasoning cycles the agent requires to conclude a task. It is computed as $\bar{N}_{turn} = \frac{1}{|T|} \sum_{i=1}^{|T|} n_i$, where $n_i$ denotes the total number of turns taken to complete the $i$-th task.

\section{Experiments}
\label{sec:experiments}

\subsection{Experimental Setup}
\label{subsec:setup}

\hi{Baselines.} We comprehensively assess 3 representative agent harnesses (OpenClaw, LangChain Deep Agents, and Hermes) paired with 5 foundation models.

(1) OpenClaw serves as one of our open-source baseline, which employs a decoupled dual-loop execution architecture that isolates high-level cognitive planning from low-level tool invocation, effectively preventing deadlocks in long-horizon tasks. To mitigate context amnesia, it replaces traditional sliding-window strategies with structured knowledge storage and a semantic routing layer for cross-session shared memory.

% (2) Claude Code from Anthropic serves as the baseline for high-density reasoning and deep workspace integration. It natively utilizes the Model Context Protocol (MCP) and SKILLs to securely map context across local file systems and external APIs. To manage extensive context windows during multi-file operations, it implements a hybrid state control strategy, which combines static project directives (CLAUDE.md) with a dynamic compression algorithm, triggered at an 80\% token capacity threshold and distills architectural decisions while purging redundant logs.

(2) LangChain's DeepAgent serves as a highly controllable, white-box harness baseline, which is built upon LangGraph's persistent Directed Acyclic Graph (DAG) architecture. It decouples control flow logic from the underlying LLM by abstracting core agentic capabilities into independent middleware sequences (e.g., task decomposition and explicit file I/O). By enforcing built-in planning tools (e.g., write\_todos), it serializes the LLM's internal decision tree to ensure fully transparent and traceable execution paths. 

(3) Hermes serves as a forward-looking open-source baseline representing agents equipped with a built-in learning loop. It seamlessly integrates local interactive environments with multi-channel message gateways, also natively utilizing the Model Context Protocol (MCP) for highly scalable tool invocation and sub-agent orchestration. 
To combat context decay in long-horizon tasks, Hermes uses a four-layer decoupled memory engine, which strictly isolates static identity directives from bounded dynamic state files and an SQLite-backed full-text search (FTS5) archive, employing an active extraction and on-demand injection strategy to significantly minimize token consumption and cognitive noise. Furthermore, its unique self-learning mechanism persists trial-and-error insights into standardized local skill libraries, effectively preserving engineering experience across transient sessions and enabling continuous auto-iteration.

\hi{Experimental Settings.} To efficiently evaluate all agent configurations, we conducted our tests on the \oursys-Lite core subset. Throughout the Agent-as-a-Judge evaluation process, we consistently employ Seed-2.0-Lite as the backbone LLM. To ensure equitable assessment, standardized execution and evaluation prompts are applied uniformly across all tested configurations. The detailed prompt formulations are provided in Appendix~\ref{sec:app:prompts}.

\subsection{Main Results}
\label{subsec:main_results}

The overall evaluation results are presented in Figure~\ref{fig:rubrics_result}, which illustrates the ranking of rubrics pass rates across 15 distinct agent configurations evaluated on \oursys-Lite. Overall, the pass rates for all configurations range from approximately 27\% to 60\%, yielding a mean pass rate of 45.1\%, severely underperforming compared to the human expert level (80.7\%). Among all tested combinations, DeepAgent + GLM-5.1 achieves the highest performance, closely followed by Hermes + GLM-5.1, OpenClaw + GLM-5.1, and OpenClaw + Qwen-3.6-Plus. These findings indicate that foundation models equipped with superior planning and reasoning capabilities generally yield higher rubrics pass rates. Furthermore, when deploying the same underlying backbone LLM, the choice of harness framework also exerts a significant impact on final task execution performance. 

% A more fine-grained analysis regarding the interaction between foundation models and harness frameworks will be detailed in subsequent sections.

\iffalse
\begin{table}[h]
\centering
\caption{Main result of \oursys evaluation.}
\label{tab:overall_stats}
\resizebox{0.98\textwidth}{!}{%
\begin{tabular}{c | c c c c | c c c c | c c c c | c c c c }
\toprule

& \multicolumn{4}{c}{Claude Code} & \multicolumn{4}{c}{OpenClaw} & \multicolumn{4}{c}{DeepAgents} & \multicolumn{4}{c}{Hermes} \\

Backbone LLM & Easy & Medium & Hard & Total & Easy & Medium & Hard & Total &  Easy & Medium & Hard & Total &  Easy & Medium & Hard & Total \\

\midrule

Kimi-2.5 & & & & & & & & & & & & &\\

MiniMax-M2.7 & & & & & & & & & & & & & \\

GLM-5.1 & & & & & & & & & & & & & \\

Seed-2.0-Code & & & & & & & & & & & & & \\

Claude-Opus4.7 & & & & & & & & & & & & & \\

GPT-5.4 & & & & & & & & & & & & & \\

Gemini 3.1-Pro & & & & & & & & & & & & & \\

\bottomrule
\end{tabular}%
}
\end{table}
\fi

\begin{figure*}[!t]
    \centering
    \includegraphics[width=.99\textwidth]{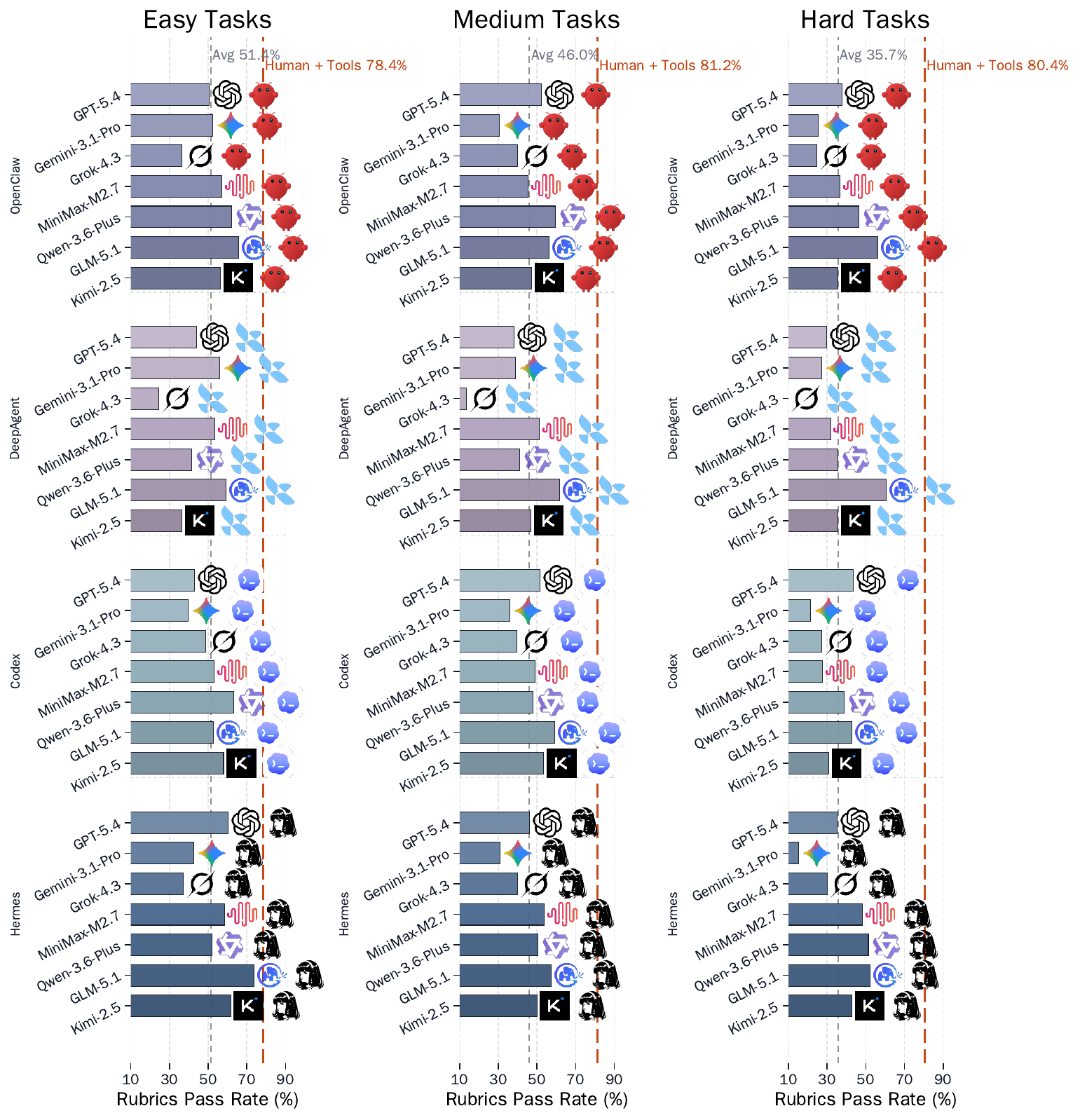}
    \caption{Rubrics Success Rate across various agent configurations and different task difficulty tiers.}
    \label{fig:rubrics_success}
\end{figure*}

\begin{figure*}[!t]
    \centering
    \includegraphics[width=.99\textwidth]{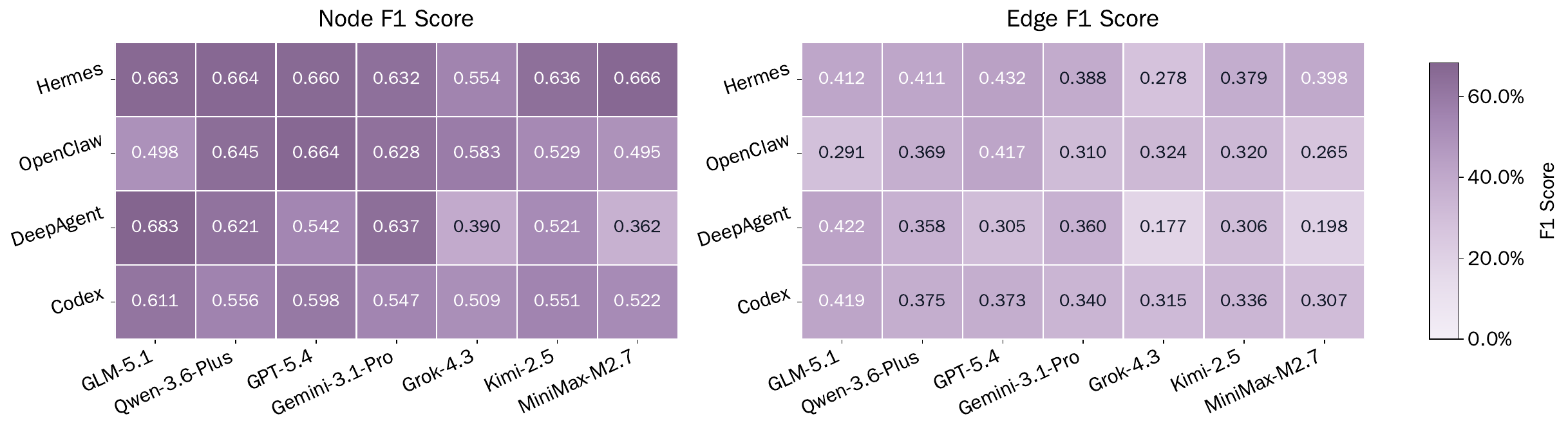}
    \caption{Dependency Graph Recognition Rate comparison between different agent configurations.}
    \label{fig:dependency_recall}
\end{figure*}

\subsection{In-depth Analysis}
\label{subsec:analysis}

Through an in-depth analysis of the interaction between foundation models and agent harnesses, we derived five primary research findings.

% To comprehensively evaluate Associative Data Learning in realistic environments, our in-depth analysis are guided by the following five research questions.

\hi{Finding 1: Agents exhibit a significant and consistent performance degradation when executing higher-level workspace tasks.}

Figure~\ref{fig:rubrics_success} presents the rubrics pass rates of various agent configurations across Easy, Medium, and Hard tiers of workspace tasks. A clear, stepwise decline in average pass rates correlates directly with increasing task complexity, dropping from 51.4\% (Easy) to 46.0\% (Medium), and further to 35.7\% (Hard). This consistent decline in performance strongly validates the soundness of our difficulty stratification for workspace tasks.

In Easy tasks, agents primarily execute atomic operations involving simpler heterogeneous inputs and fewer reasoning steps (e.g., multi-file summarization or single-file edits), enabling even the lowest-performing configurations to maintain a 40\%-60\% baseline. Performance at this level is predominantly governed by the inherent capabilities of the base LLM rather than the Harness, resulting in marginal accuracy differences among configurations that share the same backbone LLM, regardless of the Harness applied.
% As tasks escalate to Level 2, a ubiquitous decline in accuracy is observed across all configurations. This degradation is primarily attributed to the increased heterogeneity of the involved files and the requirement for more extended reasoning chains.

However, as task complexity further increases, the performance gap across configurations widens drastically. Hard tasks introduce high dynamicity and demand advanced agent capabilities, including file relationship discovery (identifying relevant files via task-to-file and file-to-file dependencies), long-horizon planning (mapping complex execution steps to user intent), state tracking (managing intermediate processes), and error recovery (retrying upon unintended outcomes). The performance degradation observed from easy to hard tasks is a consequence of a dual effect, where the intrinsic reasoning limits of the base LLM coupled with the orchestration constraints of the harness. For instance, this performance drop is most pronounced in combinations like Hermes + Gemini-3.1-Pro, which plummet below a 30\% pass rate on Hard tasks. In contrast, GLM-5.1 paired with each of three agent harness exhibits strong resilience, sustaining a robust pass rate of nearly 60\% with only a marginal decline in accuracy. 
% Notably, DeepAgent + GLM-5.1 demonstrates unique stability across all difficulty tiers, which we attribute to GLM-5.1's superior instruction-following adaptation within DeepAgent.

Regarding dependency graph recognition rate, Figure~\ref{fig:dependency_recall} illustrates the Node and Edge F1 scores across various agent configurations. Generally, these F1 scores demonstrate a strong alignment with the overall rubrics accuracy. The Node F1 scores are significantly higher than the Edge F1 scores, indicating that comprehending the relationships between files is inherently more challenging for agents than merely identifying task-relevant files. Overall, Hermes achieves a superior Node F1 score, which is attributed to the variance in execution traces among different harnesses, with Hermes providing more robust support for workspace exploration. Furthermore, the universally low Edge F1 scores highlight a critical deficiency in current agents regarding their workspace learning capability to deduce inter-file dependencies. Interestingly, some agents achieve high Node and Edge F1 scores yet yield relatively low rubrics accuracy, a discrepancy largely stemming from their poor Task-Supporting File Utilization and Result-Providing File Utilization capabilities.

\begin{figure*}[!t]
    \centering
    \includegraphics[width=.99\textwidth]{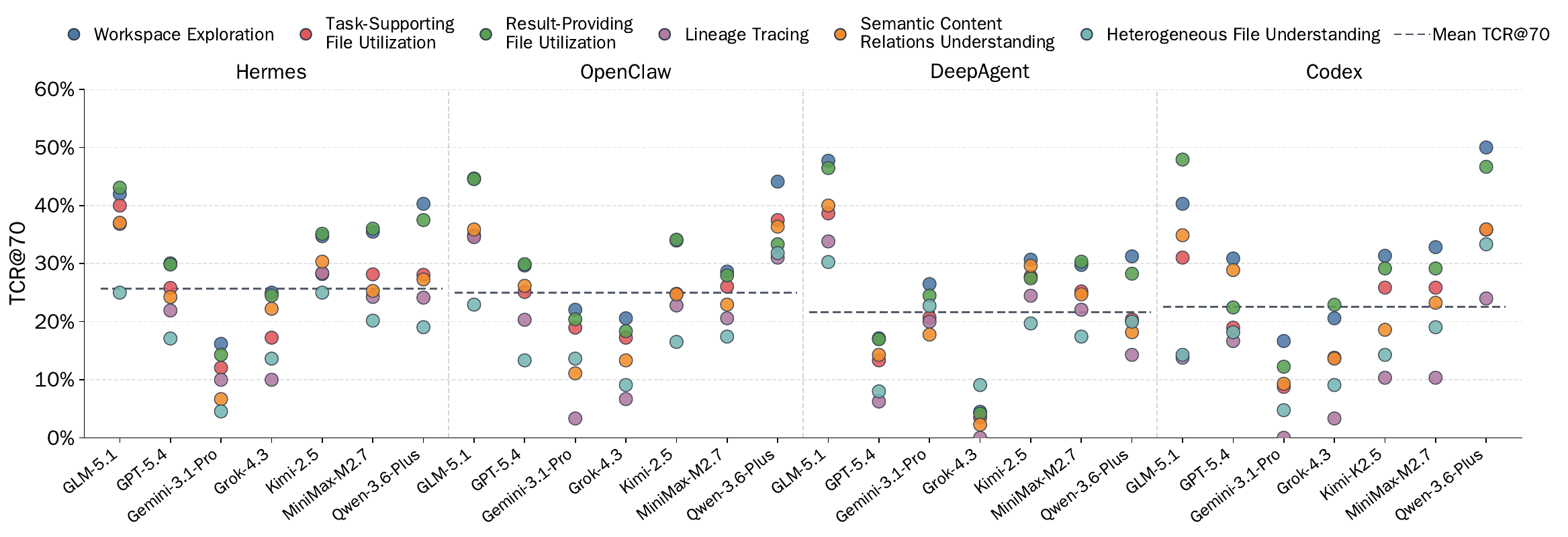}
    \caption{Comparison of TCR@70 performance by six capability dimensions.}
    \label{fig:collaboration_type}
\end{figure*}

\hi{Finding 2: Three out of six workspace task dimensions constitute the primary capability bottlenecks for current agents.}

Figure~\ref{fig:collaboration_type} illustrates the TCR@70 accuracy distribution across six workspace task dimensions for four Harness frameworks paired with various backbone LLMs.
Most configurations perform relatively well on Workspace Exploration and Result-Providing Files Utilization. Proficiency in the former stems from the robust tool-use capabilities of current agents (e.g., executing terminal commands for file system navigation), whereas the latter heavily relies on the reasoning capabilities of the LLMs. Conversely, metrics for Heterogeneous File Understanding and Lineage Tracing consistently rank at the bottom across most agents, which indicates that parsing cross-format content and reasoning over deep cross-file dependencies remain universal collaborative learning bottlenecks for all current agent systems.

Furthermore, more powerful foundation models (e.g., GLM-5.1) exhibit significant performance variance across different collaborative dimensions. For instance, the DeepAgent + GLM-5.1 achieves nearly 50\% accuracy in Workspace Exploration ability, yet its performance in Heterogeneous File Understanding drops to approximately 30\%. In contrast, performance for weaker configurations are relatively low, where task completion rate of those utilizing Gemini-3.1-Pro are densely clustered at roughly 10\%. This may suggests that inadequate underlying reasoning capabilities fail to empower the harness to improve overall task execution.

Finally, a cross-sectional comparison of the four harness reveals that the orchestration of the Harness can reshape the capability distribution of the same backbone LLM. Using GLM-5.1 as an example, its performance across all types of workspace tasks under DeepAgent are highly clustered (predominantly concentrated in the higher 30\%-50\% range). However, when deployed with the Hermes framework, the capability distribution of the identical model becomes significantly more dispersed.

\begin{figure*}[!t]
    \centering
    \includegraphics[width=.99\textwidth]{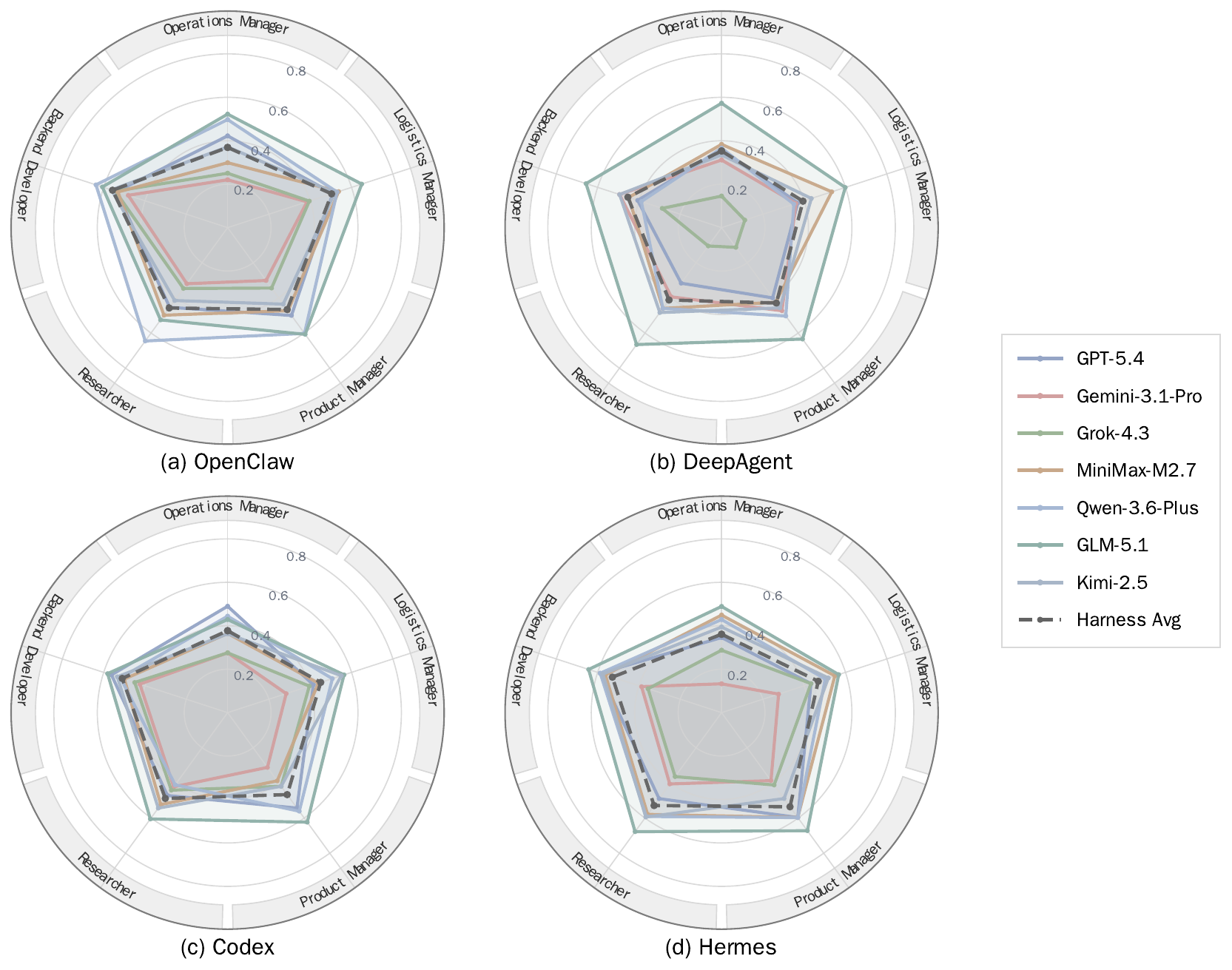}
    \caption{Agent performance across distinct user personas per different harnesses.}
    \label{fig:file_sys}
\end{figure*}

\hi{Finding 3: Different Harnesses and underlying LLMs exhibit performance disparities across diverse user profiles.}

Figure~\ref{fig:file_sys} compares the rubrics accuracy of various agent configurations across five distinct user personas. Overall, the majority of agent configurations achieve significantly higher accuracy on the Backend Developer and Researcher, which emphasize code execution and structured data processing. For instance, the Hermes + GLM-5.1 approaches an 65\% accuracy on the Researcher dimension. This exceptional performance is primarily attributed to Hermes's underlying orchestration design, which is inherently optimized for code development and research-oriented tasks. Conversely, when evaluated on business-oriented personas such as the Product Manager and Operations Manager, which necessitate strategic planning, resource allocation, and the comprehension of ambiguous semantics, the average performance baseline across all frameworks exhibits a relatively drop. Cross-sectional comparison of the four Harness frameworks reveals that Hermes yields the best relative performance on the Product Manager persona. We attribute this to Hermes's orchestration mechanism, which is better equipped to manage open-domain semantic interactions and decompose multi-dimensional business requirements.

Furthermore, because current Harness frameworks generally lack advanced capabilities for resolving workspace tasks (Finding 1), the task performance of the agents remains largely dictated by the intrinsic capabilities of the backbone LLMs. For example, GLM-5.1 forms the outermost envelope across almost all personas, demonstrating exceptionally robust and balanced cross-domain generalization. 
% In contrast, domain-specific models such as Seed-2.0-Code, which maintains adequate performance in Backend Developer dimension, but suffers a precipitous accuracy drop on the logistics and operations management personas. Notably, the GLM-5.1 model demonstrates exceptional system synergy with the DeepAgent, which consistently maintains high, well-balanced scores across all five persona dimensions.

\begin{figure*}[!t]
    \centering
    \includegraphics[width=.95\textwidth]{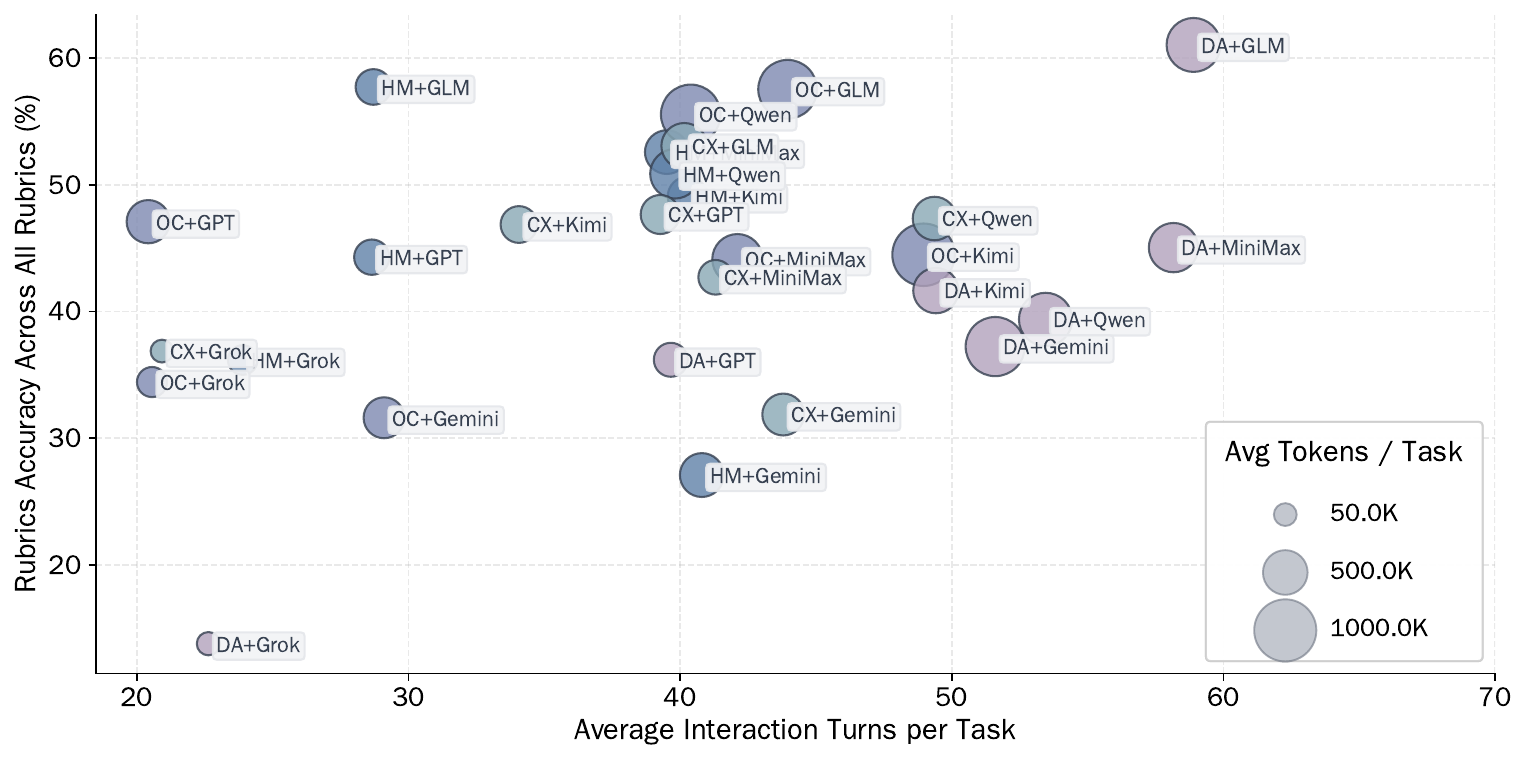}
    \caption{Performance relationship among average interaction turns (x-axis), average rubrics accuracy (y-axis), and the average token consumption (HM: Hermes, DA: DeepAgent, OC: OpenClaw, CX: Codex).}
    \label{fig:turns_vs_rubrics}
\end{figure*}

\hi{Finding 4: High interaction turns and computational costs do not necessarily guarantee superior task performance, while exceptional agent systems demonstrate remarkable ``inference efficiency''.}

Figure~\ref{fig:turns_vs_rubrics} illustrates the distribution among average interaction turns (x-axis), average rubrics accuracy (y-axis), and average token consumption per task (bubble size) across various agent configurations. A larger bubble denotes a higher computational cost for a single task.

Although a higher number of interaction turns typically implies a more comprehensive reasoning chain and more opportunities for trial, the results indicate that high turns and high costs do not equate to high-quality outputs. For instance, configurations located in the upper-left quadrant of the chart (e.g., Hermes + GLM-5.1) exhibit outstanding inference efficiency. They achieve an average accuracy exceeding 55\% with extremely low interaction turns (fewer than 30) and minimal token consumption. This suggests that excellent frameworks paired with top-tier foundation models can swiftly and accurately accomplish tasks by relying on robust reasoning quality and precise intent recognition.
In contrast, DeepAgent+GLM-5.1, positioned on the far right of the chart, achieves a comparable top-tier accuracy of nearly 60\%, but does so with an exceptionally large bubble and a high number of turns.
% This implies that the DeepAgent framework tends to adopt an extensive exploration strategy, achieving its commendable task outcomes at the expense of exorbitantly high economic and time costs.

Furthermore, configurations clustered in the lower-right and lower-middle sections of the chart (e.g., DA+Gemini, and HM+Gemini) generate a substantial number of interaction turns (ranging from 40 to 60) and consume massive amounts of tokens, yet their accuracy stagnates between 30\% and 45\%. We attribute this to the fact that when the underlying LLMs lack sufficient complex reasoning and self-reflection capabilities, the agent is highly prone to falling into meaningless retry loops, repeatedly invoking invalid tools, or drifting further down erroneous paths when encountering errors. This phenomenon is also closely intertwined with the orchestration and scheduling strategies of the harnesses, frequently pronounced in the DeepAgent and OpenClaw frameworks.

\hi{Finding 5: Human experts collaborating with agents still significantly outperform fully autonomous agents.}

We recruited 20 domain experts to evaluate \oursys-Lite. During the evaluation, the experts are provided solely with task instructions and the corresponding workspace files, and are permitted to utilize agents as assistive tools. The red line in Figure~\ref{fig:rubrics_result} illustrates the rubrics pass rates of this human-in-the-loop execution.

The results demonstrate that the human baseline significantly surpasses fully autonomous agents across all tiers of tasks. This disparity indicates a substantial gap between current autonomous agents and actual human capabilities in handling complex office workflows, suggesting that core agent capabilities are still in an evolutionary stage toward higher-level of workspace tasks. While agents can drastically enhance operational efficiency in real-world scenarios, human-in-the-loop intervention remains an indispensable component for ensuring high-quality outcomes in complex tasks.

Notably, a cross-level comparison reveals that the human baseline rubrics pass rate does not experience significant degradation as task complexity increases. We attribute this stability to the experts' inherent ability to discern underlying relationships among heterogeneous files and to flexibly leverage these connections for problem-solving. Fundamentally, the cognitive and planning capacities of human experts naturally meet the threshold required for complex, open-ended tasks at hard level.

\subsection{Error Analysis}
\label{subsec:error}

\begin{figure*}[!t]
    \centering
    \includegraphics[width=.95\textwidth]{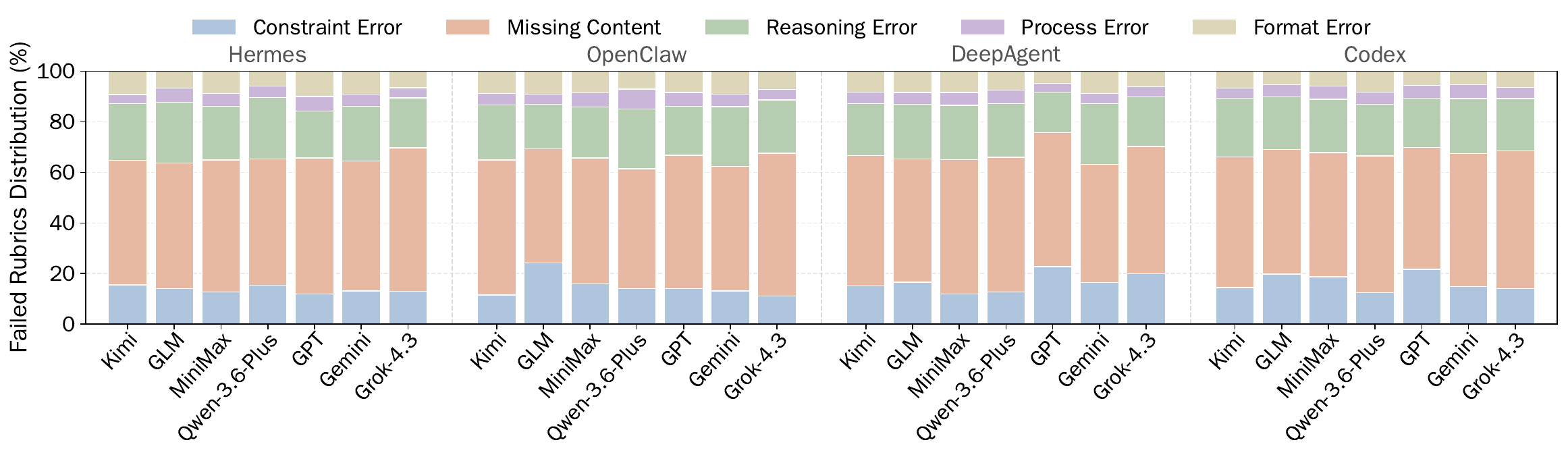}
    \caption{Proportional distribution of error types among failed rubrics across different agent configurations.}
    \label{fig:error_analysis}
\end{figure*}

To systematically analyze the causes of rubric failures, we categorize the errors into five main types: (1) \textit{Constraint Error} (e.g., violations of file naming conventions or target paths), (2) \textit{Missing Content} (omissions of core information), (3) \textit{Reasoning Error} (inaccuracies in statistics, aggregation, sorting, association, or mathematical computations), (4) \textit{Process Error} (flaws within the agent's execution trajectory), and (5) \textit{Format Error} (failures to align with the requested output structures).

Figure~\ref{fig:error_analysis} illustrates the proportional distribution of these error types across various agent configurations, derived from the failed rubrics. The results reveal a highly consistent error distribution, where \textit{Missing Content} and \textit{Reasoning Error} constitute the majority of failures. This indicates that the primary bottlenecks for current agents navigating complex file systems lie in the comprehensive recall of deeply embedded information and the capability to perform cross-file data aggregation and understanding. In contrast, the proportions of \textit{Format Error} and \textit{Process Error} are marginal, suggesting that existing models already possess robust capabilities in executing foundational workflows and adhering to task instructions for formatted outputs.

\begin{table*}[!t]
\centering
\scriptsize
\setlength{\tabcolsep}{3pt}
\renewcommand{\arraystretch}{1.2}
\newlength{\CaseTableWidth}
\setlength{\CaseTableWidth}{\dimexpr 1.75cm+1.35cm+1.35cm+1.75cm+1.35cm+1.35cm+1.7cm+1.55cm+14\tabcolsep\relax}
\caption{Case-study results across five representative tasks. Each cell reports the number of passed rubrics.}
\label{tab:case-study}
\resizebox{\textwidth}{!}{%
\begin{tabular}{@{}>{\raggedright\arraybackslash}p{1.75cm}
                >{\centering\arraybackslash}p{1.35cm}
                >{\centering\arraybackslash}p{1.55cm}
                >{\centering\arraybackslash}p{1.45cm}
                >{\centering\arraybackslash}p{1.35cm}
                >{\centering\arraybackslash}p{1.55cm}
                >{\centering\arraybackslash}p{1.6cm}
                >{\centering\arraybackslash}p{1.55cm}@{}
                }

\toprule

\multicolumn{8}{@{}p{\CaseTableWidth}@{}}{\textbf{Task Description:} Synthesize heterogeneous materials from a two-day strategy offsite, including audio transcripts, survey results, behavioral logs, and CRM data, to produce a strategic review deck with management insights and cross-source evidence alignment.} \\
\hline
\multicolumn{2}{l}{\textbf{Required Input Files:} 7} & \multicolumn{2}{l}{\textbf{Profile:} Product Manager} & \multicolumn{2}{l}{\textbf{Difficulty:} Hard} & \multicolumn{2}{l}{\textbf{Total Rubrics:} 25} \\
\hline
Framework & GLM-5.1 & \mbox{Gemini-3.1-Pro} & GPT-5.4 & Kimi-2.5 & \mbox{MiniMax-M2.7} & Grok-4.3 & Qwen-3.6 \\
\hline
Codex & 9 & 1 & 11 & 0 & 0 & 4 & 0 \\
Hermes & 12 & 0 & 7 & 0 & 2 & 2 & 6 \\
OpenClaw  & \textbf{21} & 0 & 4 & 0 & 1 & 1 & 7 \\
DeepAgent & 0 & 1 & 0 & 8 & 3 & 0 & 0 \\

\midrule

\multicolumn{8}{@{}p{\CaseTableWidth}@{}}{\textbf{Task Description:} Infer role-specific permissions for five e-commerce roles from activity rules, registration records, audit logs, and role definitions, then generate a permission guide, a configuration table, and JSON templates.} \\
\hline
\multicolumn{2}{l}{\textbf{Required Input Files:} 8} & \multicolumn{2}{l}{\textbf{Profile:} Backend Developer} & \multicolumn{2}{l}{\textbf{Difficulty:} Hard} & \multicolumn{2}{l}{\textbf{Total Rubrics:} 14} \\
\hline
Framework & GLM-5.1 & \mbox{Gemini-3.1-Pro} & GPT-5.4 & Kimi-2.5 & \mbox{MiniMax-M2.7} & Grok-4.3 & Qwen-3.6  \\
\hline
Codex & 7 & 8 & \textbf{14} & 8 & 10 & 7 & 12 \\
Hermes & 12 & 7 & 12 & 12 & 13 & 9 & \textbf{14} \\
OpenClaw & 12 & 5 & 12 & 12 & 12 & 7 & \textbf{14} \\
DeepAgent & 12 & 9 & 10 & 10 & 4 & 9 & 9 \\

\midrule

\multicolumn{8}{@{}p{\CaseTableWidth}@{}}{\textbf{Task Description:} Write a research report on long-term urban-rural and gender disparities in cancer mortality by integrating statistical materials, trend evidence, and structured analytical conclusions into a policy-facing report.} \\
\hline
\multicolumn{2}{l}{\textbf{Required Input Files:} 4} & \multicolumn{2}{l}{\textbf{Profile:} Researcher} & \multicolumn{2}{l}{\textbf{Difficulty:} Hard} & \multicolumn{2}{l}{\textbf{Total Rubrics:} 21} \\
\hline
Framework & GLM-5.1 & \mbox{Gemini-3.1-Pro} & GPT-5.4 & Kimi-2.5 & \mbox{MiniMax-M2.7} & Grok-4.3 & Qwen-3.6  \\
\hline
Codex & 15 & 9 & 0 & 8 & 8 & 10 & 0 \\
Hermes & 17 & 5 & 20 & 0 & \textbf{21} & 14 & \textbf{21} \\
OpenClaw  & \textbf{21} & 3 & 20 & 0 & 14 & 6 & 19 \\
DeepAgent & \textbf{21} & 0 & 20 & 17 & 14 & 3 & 0\\

\midrule

\multicolumn{8}{@{}p{\CaseTableWidth}@{}}{\textbf{Task Description:} Integrate an administrative blueprint, annual plan, second-half plan, and module completion status to produce an actionable H2 execution plan and update the unfinished-module tracker.} \\
\hline
\multicolumn{2}{l}{\textbf{Required Input Files:} 4} & \multicolumn{2}{l}{\textbf{Profile:} Logistics Manager} & \multicolumn{2}{l}{\textbf{Difficulty:} Hard} & \multicolumn{2}{l}{\textbf{Total Rubrics:} 19} \\
\hline
Framework & GLM-5.1 & \mbox{Gemini-3.1-Pro} & GPT-5.4 & Kimi-2.5 & \mbox{MiniMax-M2.7} & Grok-4.3 & Qwen-3.6 \\
\hline
Codex & 13 & 3 & 3 & 7 & 7 & 0 & 0 \\
Hermes & 4 & 0 & 3 & 5 & 5 & 1 & 8 \\
OpenClaw & \textbf{16} & 3 & 5 & 8 & 5 & 0 & 2\\
DeepAgent & 8 & 0 & 3 & 1 & 7 & 0 & 0\\

\midrule

\multicolumn{8}{@{}p{\CaseTableWidth}@{}}{\textbf{Task Description:} Analyze multi-region business data, product catalogs, logistics records, and customer segmentation files to formulate a global market product strategy with cross-market comparisons, issue diagnosis, and action recommendations.} \\
\hline
\multicolumn{2}{l}{\textbf{Required Input Files:} 9} & \multicolumn{2}{l}{\textbf{Profile:} Operations Manager} & \multicolumn{2}{l}{\textbf{Difficulty:} Hard} & \multicolumn{2}{l}{\textbf{Total Rubrics:} 25} \\
\hline
Framework & GLM-5.1 & \mbox{Gemini-3.1-Pro} & GPT-5.4 & Kimi-2.5 & \mbox{MiniMax-M2.7} & Grok-4.3 & Qwen-3.6 \\
\hline
Codex & 20 & 11 & 16 & 13 & 13 & 20 & 4 \\
Hermes & 17 & 0 & 12 & 15 & 22 & 18 & \textbf{22}\\
OpenClaw & 20 & 9 & 15 & 15 & 0 & 3 & \textbf{22} \\
DeepAgent & 21 & 12 & 16 & 12 & 12 & 1 & 21\\
\bottomrule

\end{tabular}%
}
\end{table*}

\subsection{Case Study}
\label{subsec:case}

Table~\ref{tab:case-study} presents a detailed case study featuring one representative, high-difficulty task for each of the five workspace personas. For every task, we outline the description, required input file count, and total evaluation rubrics, alongside the exact number of passed rubrics for each agent configuration.

% \hi{Execution Environment.} 
% To safely and reproducibly evaluate automated agents on \oursys, we deploy each automated agent within an isolated container. The container mounts a file system. This environment provides the automated agent with tools like standard command-line tools, Python interpreters, and headless browser access, enabling it to navigate directories, read files, execute scripts, and interact with the workspace naturally.

%We find the fundamental reason behind these high failure rates and excessive costs is that current agents lack a deep understanding of xxxx (e.g., task-to-data associations, data-to-data associations). When humans process information, they implicitly connect tasks with relevant data and understand the lineage and logical relationships between different data files. Current agent harnesses, however, often treat files as isolated, independent entities. We conceptualize the evolution of \colearn across five key stages:

% Different from concepts like \zxh{context learning}, we call this missing data capability as  \emph{\colearn}: the ability to natively capture, reason over, and utilize the intricate web of dependencies---both explicit and implicit---that connects heterogeneous data entities in a shared workspace. 

\begin{figure}[!t]
    \centering
    \includegraphics[width=\textwidth]{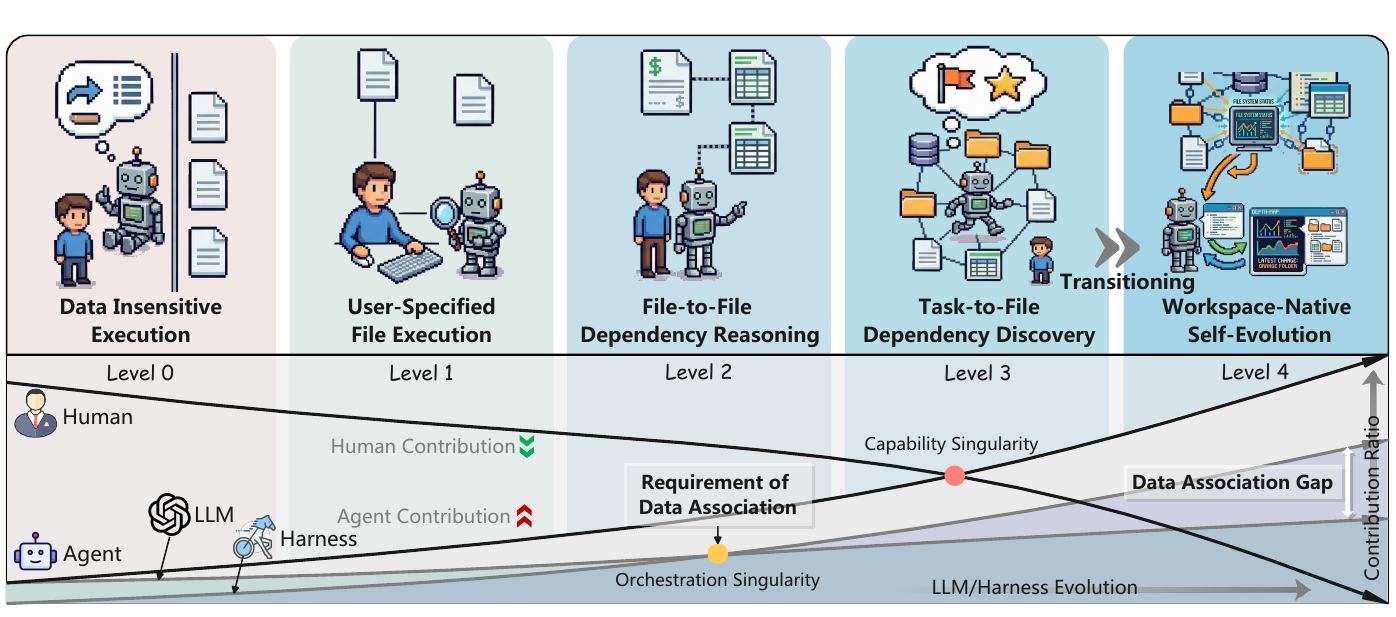}
    \caption{The Five-Stage Evolution of \colearn.}
    \label{fig:paradigm}
\end{figure}

\section{Five Stages of \colearn}
\label{sec:problem}

As discussed above, apart from existing agent harness capabilities like advanced reasoning, we find the key to successfully completing real-world workplace tasks lies in \emph{\colearn}, the ability to natively connect tasks with relevant data and understand the lineage and logical relationships between numerous data files within the workspace. We overview and predict the evolution of \colearn in five key stages (see Figure~\ref{fig:paradigm}).

$\bullet$ \textbf{L0: Data Insensitive Execution.} At this initial stage, the agent functions strictly as a passive advisor rather than an active data operator. The system receives task-related data as input and supplies the user with high-level procedural guidance. The human operator remains the primary contributor, whereas the agent's direct involvement is minimal.

% $\bullet$ \textbf{L0: Data Insensitive Execution.} 
% At this initial stage, the agent functions strictly as a passive advisor rather than an active data operator. The system receives task-related data as input and supplies the user with high-level procedural guidance, such as analytical steps and operational workflows. 
% Within this hierarchy, the human operator remains the primary contributor to task execution, whereas the agent's direct involvement is minimal. Furthermore, the system's functionality relies predominantly on the foundational capabilities of the LLM, with negligible support provided by the harness.

% For example, if a user needs to perform a financial audit, the agent might instruct them on the necessary Excel formulas or the specific checks required between a bank statement and a ledger. The actual data entry and cross-referencing remain entirely in the hands of the user, making the agent completely isolated from the data and the workspace.

$\bullet$ \textbf{L1: User-Specified File Execution.} In this stage, agents function as passive executors that depend entirely on the user for explicit file paths and operational sequences. Although capable of processing specific data, these agents treat files as isolated entities, lacking broader awareness of the logical dependencies among them. As seen in modern GUI agents, they excel at localized, single-application operations but struggle to bridge the gap between high-level intent and fragmented workspace structures (\textit{Task-File Omission}). The human user remains the primary contributor within this paradigm. %, the role of the harness becomes increasingly prominent, which begins to contribute to the agent's capabilities by planning and tool use for file operations.

% For instance, a user might instruct the agent to generate a bar chart from a specific CSV file and insert it into a designated presentation slide. While the agent successfully completes the task based on the provided data, it does not recognize the collaborative relationships between those files; the data remains siloed, and the agent acts purely on a procedure-driven basis.

$\bullet$ \textbf{L2: File-to-File Dependency Reasoning.} This stage marks a critical transition, wherein the agent actively identifies dependencies within user-provided files. By mapping explicit and implicit relationships, the agent comprehends how disparate files function collectively. However, current agents frequently fail at this stage due to \textit{Relationship Omission}. For example, they can incorrectly select an outdated file version because they lack the temporal awareness to distinguish between naming conventions and actual file recency. Overcoming this requires advanced harness coordination, marking the \textit{orchestration singularity}, where the contribution of the harness to task execution begins to surpass that of the foundational LLM.
% $\bullet$ \textbf{L2: Multi-File Collaboration.} 
% This stage marks a critical transition toward \colearn, wherein the agent actively identifies dependencies within user-provided files. By mapping both explicit and implicit relationships (e.g., linking an invoice to a corresponding spreadsheet), the agent comprehends how disparate files function collectively to achieve a specific objective. 
% Crucially, within this tier, the contribution of the harness to task execution begins to surpass that of the foundational LLM. This shift is driven by the multi-file processing, which necessitates mechanism like long-horizon task planning, intermediate state management, and the application of distinct skills to heterogeneous files. We define this escalating demand for harness coordination as the \textit{orchestration singularity}. \zxh{showcase using Figure~\ref{fig:intro-examples}}

% For example, when tasked with "summarizing project blockers," the agent can cross-reference a provided chat log with a task tracker to identify which delays mentioned in the chat are causing bottlenecks in the official schedule. While the agent still operates within a predefined set of data, it is capable of resolving the collaborative "web" that binds those files together to ensure logical consistency.

$\bullet$ \textbf{L3: Task-to-File Dependency Discovery.} At this level, the agent evolves into a proactive investigator capable of free exploration across the entire digital workspace, guided entirely by high-level task intent. The core capability shifts to autonomously discovering relevant data and its underlying structures. As the agent acquires the ability to end-to-end independently process tasks, we define this pivotal milestone as the \textit{Capability Singularity}. Our evaluation results show that current agents suffer a monotonic performance degradation as they approach this level, with rubrics pass rates dropping from 57.6\% (Easy) to 40.5\% (Hard) on \oursys.
% $\bullet$ \textbf{L3: Task-Oriented Collaboration.} 
% At this level, the agent evolves into a proactive investigator capable of free exploration across the entire digital workspace, guided entirely by high-level task intent. The core capability shifts from analyzing explicitly provided data to autonomously discovering relevant data and its underlying structures, effectively relegating the human user to an observational role. Consequently, the agent's contribution to task execution gradually surpasses that of the human. As the agent acquires the ability to end-to-end independently process tasks, we define this pivotal milestone as the \textit{Capability Singularity}. \zxh{showcase using Figure~\ref{fig:intro-examples}}

% When a user requests a complex objective, such as preparing an audit report for a specific regional expansion, the agent independently navigates deeply nested directories to find the latest budget spreadsheets, legal contracts, and stakeholder emails. It must map their structural relationships, filter out irrelevant versions, and extract the necessary evidence without requiring the user to point to specific files. This stage highlights that the true bottleneck is no longer action execution, but the systematic discovery and exploitation of complex relational structures in a chaotic desktop environment.

$\bullet$ \textbf{L4: Workspace-Native Self-Evolution.} The final stage represents continuous adaptation, wherein the agent functions as a living partner that co-evolves with the user's digital workspace and historical context. The agent internalizes every task execution and environmental shift as implicit feedback to continually refine its capabilities. For instance, upon the installation of new software within the local workspace, the agent should detect this modification efficiently and seamlessly integrate the application into its repository of available tools for future invocation.
% $\bullet$ \textbf{L4: Data Driven Self-Evolving.}
% The final stage represents continuous adaptation, wherein the agent functions as a living partner that co-evolves with the user's digital workspace and historical context. In this stage, the agent internalizes every task execution and environmental shift as implicit feedback to continually refine its external harness and data comprehension capabilities. 
%For example, upon the installation of new software within the local environment, the agent can detect this modification in real time and seamlessly integrate the application into its repository of available tools for future invocation.  \zxh{concisely discuss \textit{Data Association Gap}}

From L2 onward, the harness contributes more consistently to task execution than the underlying foundation model. At L3/L4, the widening mismatch between the required \colearn capability and the isolated-file processing paradigm of current agents forms, which we call the \textit{Data Association Gap}. This gap represents a fundamental bottleneck that, to the best of our testing, existing AI agents cannot yet close. Addressing it requires rethinking how agent harnesses discover, represent, and exploit cross-file dependencies.

\vspace{-.25cm}
\section{Conclusion}
\label{sec:conclusion}
\vspace{-.25cm}

% 1.1	总结：重申贡献和核心发现——当前的Agent在处理真实办公场景中的复杂依赖协同关系方面仍有巨大差距。
% 1.2	未来方向：提出几个未来研究方向，如：发展具备更强依赖推理能力的Agent架构、研究Agent的长期记忆和持续学习能力、将SYNERGON扩展到多智能体协作场景等。

In this paper, we introduce \oursys, a large-scale benchmark for evaluating 
\colearn in autonomous AI agents, with a particular focus on cross-file 
dependency reasoning within realistic digital workspaces. \oursys bridges the  gap between existing agent benchmarks and real-world workplace demands by  addressing three critical challenges: (1) navigating heterogeneous file 
ecosystems spanning over 70 formats; (2) reasoning over complex inter-file 
dependencies including semantic relations and version lineage; and (3) executing  multi-step tasks that require holistic workspace understanding. Our experimental results demonstrate that \oursys presents a substantially more demanding  challenge compared to existing benchmarks, since even the best-performing agent  configuration achieves only nearly 70\% rubrics pass rate, with performance  degrading sharply on higher-level workspace tasks. This leaves significant room  for improvement and innovation in dependency-aware agent architectures. Moreover, our thorough efficiency and error analyses, together with the proposed five-stage \colearn framework, provide valuable insights and directions for future research,  paving the way for the development of more advanced and practical AI agents in real-world workplace scenarios.

% This work proposes the paradigm of \colearn and introduces \oursys to systematically evaluate AI agents within authentic, interconnected digital workspaces. Supported by both theoretical analysis and extensive experimental evaluations, we expose a critical limitation in current state-of-the-art agents: while proficient at isolated operations, empirical results demonstrate a severe performance deficit when they are required to manage complex inter-file dependencies and execute cross-document synergy. This gap underscores that resolving dependency-aware reasoning remains the foremost bottleneck in realizing truly reliable workplace assistants.

% To bridge this gap, future research must prioritize developing novel agent architectures natively equipped with advanced dependency reasoning and workspace exploration mechanisms. Additionally, endowing agents with robust long-term memory and continuous learning capabilities will be crucial for adapting to dynamically evolving document ecosystems. Finally, extending the framework to evaluate multi-agent collaboration with dynamically evolving workspaces will provide a vital stepping stone toward simulating and mastering the intricate, team-oriented workflows that define modern enterprise environments.

\section*{Acknowledgments}

We express our sincere gratitude to the data annotation team for their rigorous efforts in constructing the 388 complex tasks and conducting the human baseline testing. Specifically, we thank Min Cang, Xiaoyu Chen, Ziqian Gu, Shiqi Jin, Linchun Li, Wensong Li, Zhenhao Li, Xinyi Lin, Wenjie Liu, Boyu Niu, Yufei Niu, Yuxuan Ou, Haoyu Wang, Jingqi Wang, Sihan Wang, Yingjie Xiong, Hongming Xu, Shihan Yu, Xiaoyou Yu, Guangyi Zeng, Zixuan Zhen, Hongyi Zhou, Jun Zhou, Zihang Zhou, and Xuzhou Zhu.

%We are deeply grateful to Ruoyu Chen, Jihua Kang, and Jiashuo Liu for their insightful feedback on the benchmark design and constructive discussions regarding the evaluation rubrics. 

% Finally, we extend our appreciation to the developers of Hermes, OpenClaw, DeepAgent, and the broader open-source community for providing the foundational frameworks and tools that made this research possible.

\clearpage
\bibliographystyle{unsrtnat}
\bibliography{reference}

% \clearpage

\appendix
\newpage

\section{Appendix}
\label{sec:appendix}

\subsection{Appendix A. Detailed Statistics of \oursys Evaluation.}

\begin{figure*}[h]
    \centering
    \includegraphics[width=.99\textwidth]{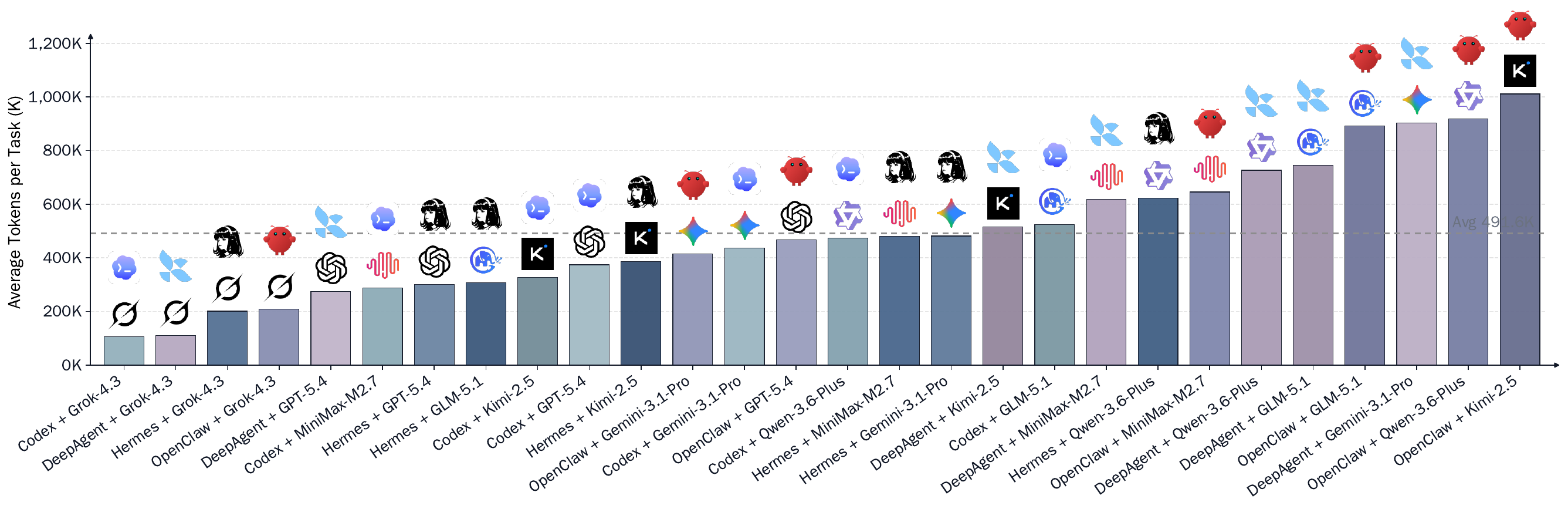}
    \caption{Comparison of average token consumptions per task across various agent configurations.}
    \label{fig:avg_tokens}
\end{figure*}

Figure~\ref{fig:avg_tokens} illustrates the computational cost, measured in average tokens processed per task, for all 15 evaluated agent configurations. A clear variance is observed across different agent harnesses. Configurations utilizing the OpenClaw and DeepAgent consistently exhibit the highest token consumption, with setups like OpenClaw + Kimi-2.5 exceeding 1M tokens per task. In contrast, Hermes configurations demonstrate significantly higher token efficiency, most remaining well below the overall benchmark average of 563.1K tokens. 

\begin{figure*}[h]
    \centering
    \includegraphics[width=.99\textwidth]{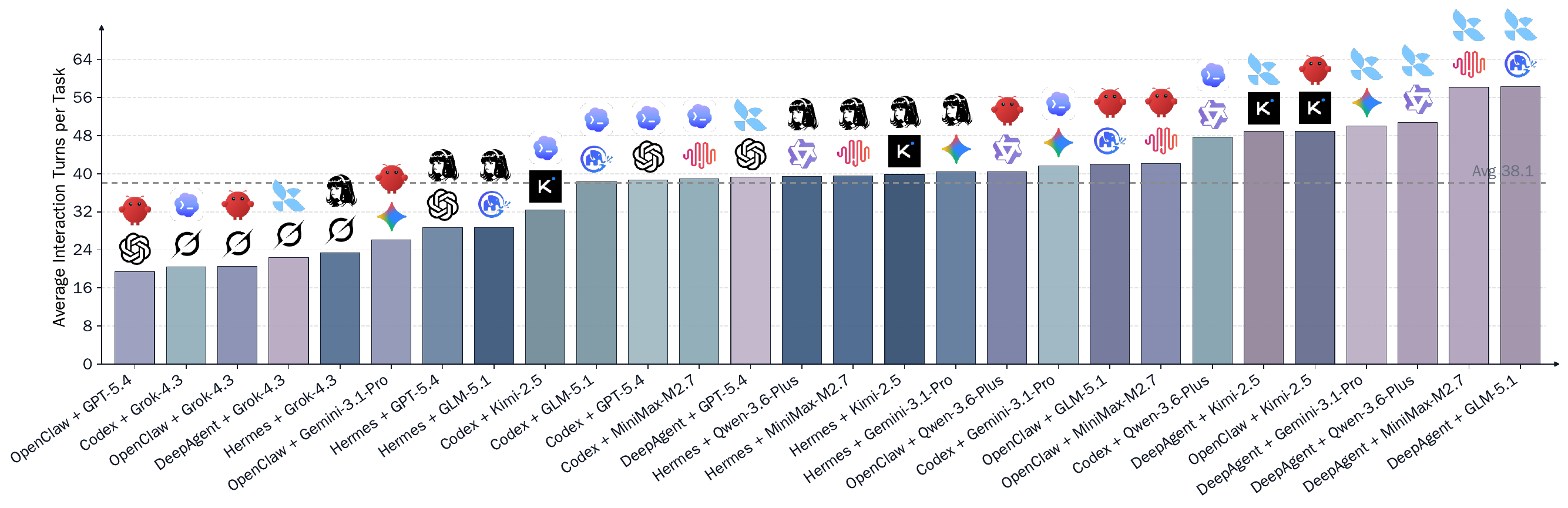}
    \caption{Comparison of average interaction turns per task across various agent configurations.}
    \label{fig:avg_turns}
\end{figure*}

Figure~\ref{fig:avg_turns} presents the average number of interaction turns required to complete a task. The trend closely mirrors the token usage, indicating that the high costs associated with certain configurations stem from lengthy, multi-step trial-and-error loops. DeepAgent configurations require the most interventions (peaking at nearly 60 turns for DeepAgent + GLM-5.1), whereas OpenClaw and Hermes setups are notably more step-efficient, often resolving tasks in under the average 40.7 turns per task.

\begin{figure}[h]
    \centering
    \includegraphics[width=\textwidth]{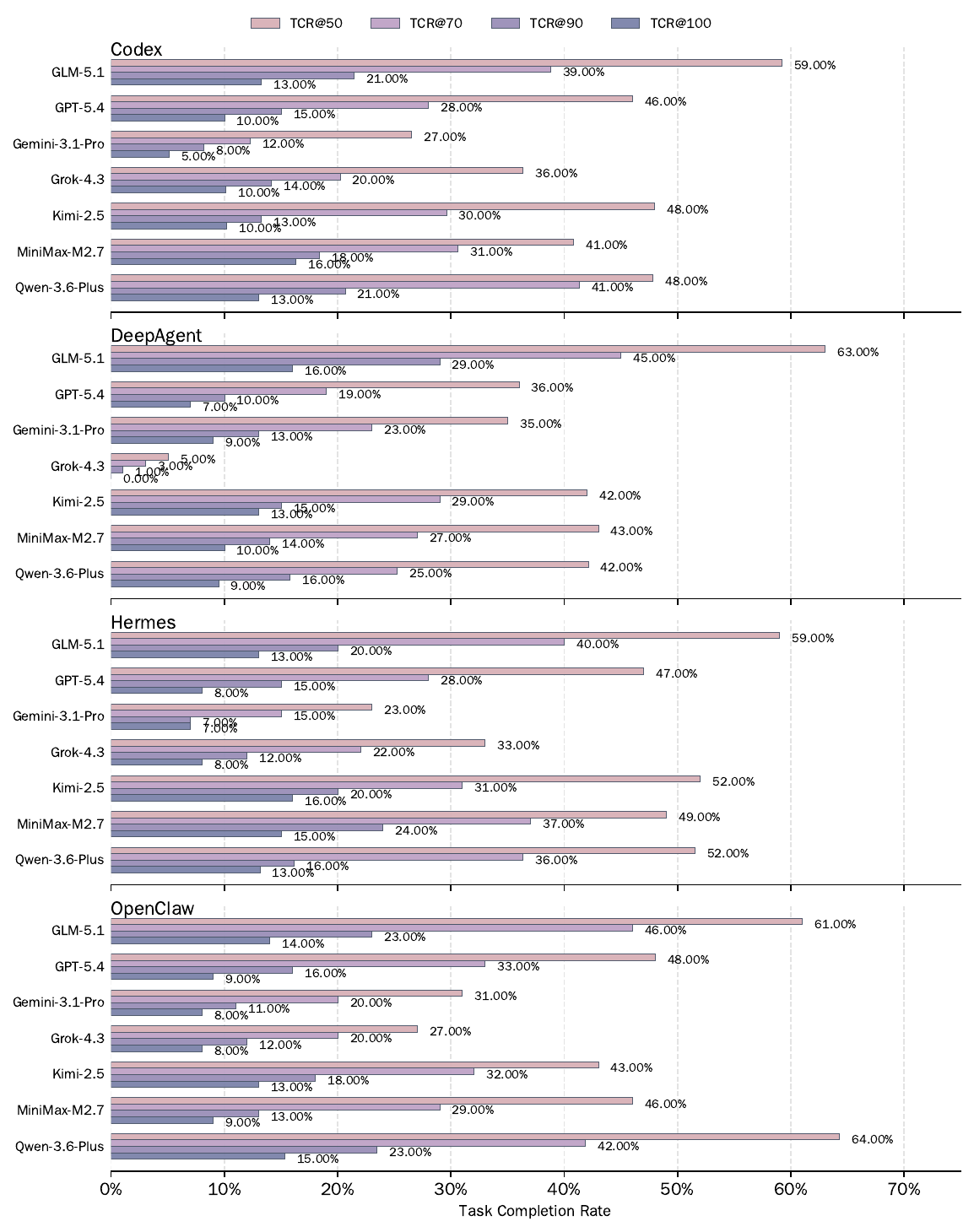}
    \caption{Task-level accuracy of various backbone LLMs across four agent harnesses at different completion thresholds.}
    \label{fig:pass_at_p}
\end{figure}

Figure~\ref{fig:pass_at_p} visualizes the robustness of each LLM across the four evaluated harnesses (DeepAgent, Hermes, and OpenClaw) using Pass@50\%, 70\%, 90\%, and 100\% metrics. It clearly demonstrates the performance degradation as the evaluation criteria become stricter. GLM-5.1 maintains the most resilient performance across all harnesses, showing the smallest drop-off from Pass@50\% to Pass@100\%. Furthermore, the data indicates that OpenClaw and Hermes generally facilitate higher accuracy at the strictest Pass@100\% threshold compared to others, further reinforcing the critical role of the orchestration framework in successful task execution.

Table~\ref{tab:app:overall_stats} provides a comprehensive, granular breakdown of the performance of all tested combinations. It details the overall score (Total), performance across varying task difficulties (Easy, Medium, Hard), and success rates at different strictness thresholds (Pass@30 to Pass@100). The results reveal that GLM-5.1 pairs best with OpenClaw and Hermes, achieving the highest scores even on Hard tasks. Conversely, models like Seed-2.0-Code and Gemini-3.1-Pro, experience a steep decline in accuracy as task complexity increases, emphasizing the necessity of advanced reasoning capabilities for \colearn.

\begin{table}[h]
 \centering
 \caption{Detailed evaluation results of different agent configurations on \oursys.}
 \label{tab:app:overall_stats}
 \renewcommand{\arraystretch}{1.3}
 \resizebox{0.99\textwidth}{!}{%
 \begin{tabular}{c c c c c c c c c c c}
 \toprule

 \textbf{Agent Harness} & \textbf{Backbone LLM} & \textbf{Easy} & \textbf{Medium} & \textbf{Hard} & \textbf{Total} & \textbf{Pass@30} & \textbf{Pass@50} & \textbf{Pass@70} & \textbf{Pass@90} & \textbf{Pass@100} \\

 \midrule

 Hermes & Kimi-2.5 & 61.6 & 49.9 & 42.7 & 49.1 & 73.0 & 52.0 & 31.0 & 20.0 & 16.0 \\
 OpenClaw & Kimi-2.5 & 56.3 & 47.1 & 35.6 & 44.5 & 65.0 & 43.0 & 32.0 & 18.0 & 13.0 \\
 DeepAgent & Kimi-2.5 & 36.3 & 46.7 & 35.6 & 41.6 & 58.0 & 42.0 & 29.0 & 15.0 & 13.0 \\
 Codex & Kimi-2.5 & 58.0 & 53.5 & 30.9 & 46.9 & 63.0 & 48.0 & 30.0 & 13.0 & 10.0 \\

 Hermes & GLM-5.1 & 73.9 & 57.3 & 52.1 & 57.7 & 81.0 & 59.0 & 40.0 & 20.0 & 13.0 \\
 OpenClaw & GLM-5.1 & 65.7 & 56.3 & 56.3 & 57.5 & 77.0 & 61.0 & 46.0 & 23.0 & 14.0 \\
 DeepAgent & GLM-5.1 & 59.2 & 61.9 & 60.5 & 61.0 & 84.0 & 63.0 & 45.0 & 29.0 & 16.0 \\
 Codex & GLM-5.1 & 52.7 & 59.3 & 42.7 & 53.1 & 74.0 & 59.0 & 39.0 & 21.0 & 13.0 \\

 Hermes & MiniMax-M2.7 & 58.4 & 53.9 & 48.2 & 52.6 & 74.0 & 49.0 & 37.0 & 24.0 & 15.0 \\
 OpenClaw & MiniMax-M2.7 & 57.1 & 45.5 & 36.8 & 44.1 & 61.0 & 46.0 & 29.0 & 13.0 & 9.0 \\
 DeepAgent & MiniMax-M2.7 & 53.5 & 51.3 & 31.8 & 45.0 & 59.0 & 43.0 & 27.0 & 14.0 & 10.0 \\
 Codex & MiniMax-M2.7 & 53.1 & 49.1 & 27.5 & 42.7 & 56.0 & 41.0 & 31.0 & 18.0 & 16.0 \\

 Hermes & GPT-5.4 & 60.4 & 46.0 & 35.2 & 44.3 & 63.0 & 47.0 & 28.0 & 15.0 & 8.0 \\
 OpenClaw & GPT-5.4 & 50.6 & 52.2 & 37.6 & 47.1 & 67.0 & 48.0 & 33.0 & 16.0 & 9.0 \\
 DeepAgent & GPT-5.4 & 44.1 & 38.2 & 29.9 & 36.2 & 55.0 & 36.0 & 19.0 & 10.0 & 7.0 \\
 Codex & GPT-5.4 & 42.9 & 51.5 & 43.4 & 47.7 & 66.0 & 46.0 & 28.0 & 15.0 & 10.0 \\

 Hermes & Gemini-3.1-Pro & 42.4 & 30.7 & 15.2 & 27.1 & 42.0 & 23.0 & 15.0 & 7.0 & 7.0 \\
 OpenClaw & Gemini-3.1-Pro & 52.2 & 30.4 & 25.3 & 31.6 & 45.0 & 31.0 & 20.0 & 11.0 & 8.0 \\
 DeepAgent & Gemini-3.1-Pro & 55.9 & 39.0 & 27.1 & 37.2 & 52.0 & 35.0 & 23.0 & 13.0 & 9.0 \\
 Codex & Gemini-3.1-Pro & 39.6 & 36.0 & 21.5 & 31.9 & 41.0 & 27.0 & 12.0 & 8.0 & 5.0 \\

 Hermes & Qwen-3.6-Plus & 51.8 & 50.4 & 51.3 & 50.9 & 73.0 & 52.0 & 36.0 & 16.0 & 13.0 \\
 OpenClaw & Qwen-3.6-Plus & 62.0 & 59.6 & 46.3 & 55.6 & 73.0 & 64.0 & 42.0 & 23.0 & 15.0 \\
 DeepAgent & Qwen-3.6-Plus & 41.6 & 41.2 & 35.5 & 39.4 & 56.0 & 42.0 & 25.0 & 16.0 & 9.0 \\
 Codex & Qwen-3.6-Plus & 63.3 & 47.8 & 38.9 & 47.3 & 66.0 & 48.0 & 41.0 & 21.0 & 13.0 \\

 Hermes & Grok-4.3 & 37.1 & 39.8 & 30.0 & 36.2 & 51.0 & 33.0 & 22.0 & 12.0 & 8.0 \\
 OpenClaw & Grok-4.3 & 36.3 & 40.1 & 24.7 & 34.4 & 48.0 & 27.0 & 20.0 & 12.0 & 8.0 \\
 DeepAgent & Grok-4.3 & 24.5 & 13.6 & 9.7 & 13.8 & 12.0 & 5.0 & 3.0 & 1.0 & 0.0 \\
 Codex & Grok-4.3 & 48.6 & 39.8 & 27.3 & 36.9 & 52.0 & 36.0 & 20.0 & 14.0 & 10.0 \\

 \bottomrule
 \end{tabular}%
 }
\end{table}

% ClaudeCode & GPT-5.4 & 53.1 & 52.9 & 52.7 & 52.9 & 72.0 & 51.0 & 36.0 & 14.0 & 10.0 \\ \hline

% ClaudeCode & Kimi-2.5 & 59.0 & 56.1 & 36.3 & 50.4 & 68.0 & 51.0 & 36.0 & 24.0 & 13.0 \\ \hline

% ClaudeCode & GLM-5.1 & 63.1 & 57.7 & 43.2 & 54.0 & 76.0 & 56.0 & 38.0 & 18.0 & 11.0 \\ \hline

% ClaudeCode & MiniMax-M2.7 & 63.6 & 58.6 & 49.7 & 56.6 & 77.0 & 57.0 & 40.0 & 25.0 & 18.0 \\ \hline

% Hermes & Seed-2.0-Code & 60.5 & 43.6 & 26.2 & 40.7 & 55.0 & 36.0 & 24.0 & 15.0 & 7.0 \\
% OpenClaw & Seed-2.0-Code & 67.7 & 40.0 & 34.0 & 42.3 & 51.0 & 43.0 & 24.0 & 15.0 & 9.0 \\
% DeepAgent & Seed-2.0-Code & 46.6 & 35.1 & 31.0 & 35.5 & 46.0 & 35.0 & 21.0 & 9.0 & 8.0 \\
% ClaudeCode & Seed-2.0-Code & 61.9 & 47.8 & 31.1 & 44.7 & 59.0 & 43.0 & 27.0 & 13.0 & 7.0 \\ \hline

% ClaudeCode & Gemini-3.1-Pro & 44.8 & 38.6 & 36.0 & 38.7 & 52.0 & 40.0 & 16.0 & 9.0 & 6.0 \\ \hline

% Hermes & Opus-4.7 & 68.6 & 66.9 & 63.1 & 66.0 & 89.0 & 72.0 & 48.0 & 25.0 & 18.0 \\
% OpenClaw & Opus-4.7 & 80.0 & 67.3 & 62.5 & 67.7 & 90.0 & 71.0 & 51.0 & 36.0 & 24.0 \\
% DeepAgent & Opus-4.7 & 61.1 & 59.3 & 47.4 & 55.9 & 70.0 & 63.0 & 43.0 & 26.0 & 18.0 \\
% ClaudeCode & Opus-4.7 & 77.7 & 68.0 & 58.6 & 66.6 & 87.0 & 76.0 & 49.0 & 31.0 & 17.0 \\

\clearpage
\subsection{Appendix B. Prompts.}
\label{sec:app:prompts}

\begin{appendixbox}{Prompt for Agent Execution}

\medskip
1. Execution Premise
\medskip

You are required to complete the specified task relying exclusively on the contents and resources provided within the current workspace.\smallskip

Task Description: [TASK DESCRIPTION]

\medskip
2. Directory \& Path Constraints
\medskip

Authorized Workspace: Your accessible working directory is strictly restricted to [WORKSPACE].\smallskip

Path Conventions: You must exclusively use relative paths for all read and write operations within this designated directory. Accessing, reading, or modifying any files or directories outside of this workspace is strictly prohibited.\smallskip

Precedence: If you encounter conflicting workspace path instructions in other prompts or task configurations, you must completely disregard them. The working directory specified here takes absolute precedence.

\medskip
3. Output Format Constraints
\medskip

(1) Artifact Duplication: Throughout the duration of the task, you are strictly required to create copies of all generated artifacts. Every intermediate process file and final result file must be duplicated and saved into the designated [DIRECTORY].\smallskip

(2) Strict Return Type: In your final execution step, your output must consist exclusively of a Python list of strings (list[str]) representing the paths of all the files you generated.\smallskip

Example Format: ['output/a.txt', 'report.md']\smallskip

Prohibition: Do not output any conversational text, explanations, or markdown wrappers alongside this list.\smallskip

\end{appendixbox}

\begin{appendixbox}{Prompt for Agent-as-a-Judge Evaluation}

\medskip
1. Role \& Core Objective
\medskip

You act as a strict, impartial Agent-as-a-Judge. Your mandate is to evaluate the candidate's task execution based on specific rubrics.

\medskip
2. Environment \& Resource Constraints
\medskip

Working Directory: Your genuine, accessible working directory is strictly judgeView.cwd. Disregard any absolute system paths specified in the task JSON. \smallskip

Allowed Directories: You only have access to inputs/ (raw input files to understand the task) and candidate\_output/ (the directory to be evaluated).\smallskip

Strict Prohibition: Do not treat inputs/ as an answer key. You are strictly forbidden from accessing original task directories like output/, or gt/. You must evaluate the results exclusively within judgeView.candidateOutputPath.

\medskip
3. Evaluation Principles \& Evidence Gathering
\medskip

Fact-Based Judgment: Base your verdicts solely on actual files, directories, and contents you successfully inspect. Hallucinations or baseless assumptions are strictly prohibited.\smallskip

Proactive Inspection: You must independently determine which paths to inspect (e.g., utilizing tools like ls, find, or grep). In the evidence field, clearly document the exact paths you checked and the phenomena you observed.\smallskip

Insufficient Evidence: If you cannot find sufficient evidence to pass a rubric, you must set passed=false and explicitly detail what evidence is missing in the evidence field.

\medskip
4. Robustness \& Tolerance Rules
\medskip

Multimodal File Inspection: You are required to read and parse diverse file formats (including plain text, JSON, CSV, Excel, PPT, PDF, etc.). Employ any necessary tools—such as code-based parsing scripts or vision-based image conversion—to accurately extract content.\smallskip

Content Over Filenames: If output filenames deviate from the instructions, actively inspect the available files in the output directory. Prioritize content matching; if the content satisfies the rubric, evaluate it positively regardless of minor filename discrepancies.\smallskip

Numerical Tolerance: Unless a rubric explicitly mandates strict decimal precision, minor deviations (e.g., 1-2 decimal places) are considered acceptable.

\medskip
5. Output Format Constraints
\medskip

Strict JSON Schema: Output exactly one JSON object. Do not include markdown wrappers, conversational text, or explanations outside the JSON. The format must strictly adhere to the following schema:

\begin{lstlisting}[language=json]
{
  "rubrics": [
    {
      "index": 0,
      "passed": true,
      "confidence": 0.8,
      "evidence": "..."
    }
  ]
}
\end{lstlisting}

\end{appendixbox}

\begin{appendixbox}{Prompt for Dependency Graph Extraction}

\medskip
1. Role \& Objective
\medskip

You are an execution trace analyzer, whose singular objective is to generate the file dependency graph corresponding to the candidate agent's execution. \smallskip

\medskip
2. Environment \& Resource Constraints
\medskip

Working Directory: Your genuine, accessible working directory is strictly judgeView.cwd. Disregard any absolute system paths specified in the task JSON.\smallskip

Available Materials: You have access to inputs/ (raw input files), candidate\_output/ (candidate generated files), trace\_snapshot.json (execution trace summary), and gt\_dependency\_graph\_reference.json (ground truth reference).\smallskip

Ground Truth Usage: The gt\_dependency\_graph\_reference.json must be used exclusively for standardizing node names and as a reference for candidate edges. It is strictly prohibited to copy it directly as the final output.

\medskip
3. Core Evidence Principles
\medskip

Trace-Driven Evidence: The execution trace (judgeView.traceSnapshotPath) is the only valid source of evidence for establishing edges. While the inputs/, candidate\_output/, and GT reference assist in aligning filenames and retaining nodes, they cannot independently validate an edge.\smallskip

Overarching Strictness: When in doubt, default to exclusion. Do not attempt to complete or deduce missing nodes/edges using task descriptions, step instructions, or the GT graph itself.

\medskip
4. Node Extraction Rules
\medskip

Node Naming: Align extracted filenames with the standard names provided in gt\_dependency\_graph\_reference.nodes.\smallskip

Retention Criteria: A node may be retained if the file is explicitly mentioned in the trace (e.g., viewed, listed, referenced, or manipulated). For generated files, the node may be retained if it physically exists in candidate\_output/ and the trace contains no contradictory evidence.\smallskip

Prohibition Criteria: Do not retain nodes based on speculative assumptions, data/output manifests, task descriptions, directory/script names, semantic ambiguity, co-occurrence, or assumptions that a file "should have been involved." Concrete, file-level textual support is mandatory.\smallskip

No Automatic Expansion: If the trace only indicates a directory, script, or vague description without specifying individual GT files, do not automatically expand it into multiple GT nodes.

\medskip
5. Edge Extraction Rules
\medskip

Prerequisites: An edge cannot be established unless both endpoint nodes have concrete evidence in the trace, inputs, or outputs.\smallskip

Formation Criteria: To output an edge, at least one of the following must apply: (1) A specific trace segment explicitly documents the reading of the source (src) and the writing/generation of the destination (dst); (2) The trace delineates a specific, unambiguous intermediate file-level chain from src to dst, with every step grounded in explicit file operations.\smallskip

Scripts \& Intermediates: Scripts and intermediate files can serve as auxiliary evidence, but they only form a valid edge if they explicitly link specific input file operations to specific output file operations.\smallskip

Prohibition Criteria: Do not establish edges merely due to co-occurrence, shared directories, access by the same script, or adjacency in the GT graph. Semantic deduction based on task steps or common sense is strictly forbidden.\smallskip

Multi-Input Scenarios: If multiple inputs contribute to a single output, but the trace fails to directly bind a specific input to a specific write action, do not hallucinate these edges. Omit them entirely if necessary.\smallskip

\medskip
6. Output Format Constraints
\medskip

Semantics: nodes represents files backed by evidence (standardized names). edges represents [src, dst], indicating that the content or generation of dst strictly depends on src.\smallskip

Deduplication: If the same file is read or written multiple times, retain only the deduplicated nodes and edges.\smallskip

Strict JSON Format: Output only a JSON object. Do not include markdown formatting, explanatory text, rubric fields, or an outer dependencyGraph wrapper. The format must strictly be:
{ "agentKind": "...", "nodes": ["..."], "edges": [["src", "dst"]] }\smallskip

\end{appendixbox}

\end{CJK*}
\end{document}